\documentclass{article}

\usepackage{microtype}
\usepackage{graphicx}
\usepackage{subcaption}
\usepackage{booktabs} 

\usepackage{hyperref}

\usepackage[preprint]{icml2026}

\usepackage{amsmath}
\usepackage{amssymb}
\usepackage{mathtools}
\usepackage{amsthm}

\usepackage[capitalize,noabbrev]{cleveref}

\theoremstyle{plain}

\theoremstyle{definition}

\theoremstyle{remark}

\usepackage[textsize=tiny]{todonotes}

\icmltitlerunning{DDP-WM: Disentangled Dynamics Prediction for Efficient World Models}

\begin{document}

\twocolumn[
  \icmltitle{DDP-WM: Disentangled Dynamics Prediction for Efficient World Models}



  \icmlsetsymbol{equal}{*}

  \begin{icmlauthorlist}
    \icmlauthor{Shicheng Yin}{equal,yyy}
    \icmlauthor{Kaixuan Yin}{equal,yyy}
    \icmlauthor{Weixing Chen}{yyy}
    \icmlauthor{Yang Liu}{yyy,zzz}
    \icmlauthor{Guanbin Li}{yyy}
    \icmlauthor{Liang Lin}{yyy,zzz}
  \end{icmlauthorlist}

  \icmlaffiliation{yyy}{School of Computer Science and Engineering, Sun Yat-sen University, Guangzhou, China}
   \icmlaffiliation{zzz}{X-Era AI Lab}
   \icmlcorrespondingauthor{Yang Liu}{liuy856@mail.sysu.edu.cn}

  \icmlkeywords{Machine Learning, ICML}

  \vskip 0.3in
]



\printAffiliationsAndNotice{}  

\begin{abstract}


World models are essential for autonomous robotic planning. However, the substantial computational overhead of existing dense Transformer-based models significantly hinders real-time deployment. To address this efficiency-performance bottleneck, we introduce DDP-WM, a novel world model centered on the principle of Disentangled Dynamics Prediction (DDP).
We hypothesize that latent state evolution in observed scenes is heterogeneous and can be decomposed into sparse primary dynamics driven by physical interactions and secondary context-driven background updates. DDP-WM realizes this decomposition through an architecture that integrates efficient historical processing with dynamic localization to isolate primary dynamics. By employing a cross-attention mechanism for background updates, the framework optimizes resource allocation and provides a smooth optimization landscape for planners.
Extensive experiments demonstrate that DDP-WM achieves significant efficiency and performance across diverse tasks, including navigation, precise tabletop manipulation, and complex deformable or multi-body interactions. Specifically, on the challenging Push-T task, DDP-WM achieves an approximately $9\times$ inference speedup and improves the MPC success rate from $90\%$ to $98\%$ compared to state-of-the-art dense models. The results establish a promising path for developing efficient, high-fidelity world models. Codes is available at \url{https://hcplab-sysu.github.io/DDP-WM/}.

\end{abstract}

\section{Introduction}
\label{chap:introduction}

\begin{figure}[t]
    \centering
    \begin{subfigure}{\columnwidth} 
        \centering
        \includegraphics[width=0.95\textwidth]{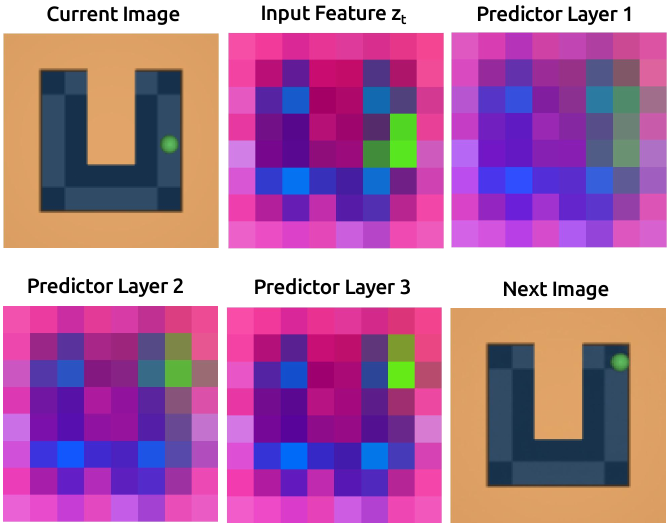}
        \caption{}
        \label{fig:intro_fig1_a}
    \end{subfigure}
    \vspace{1em} 
    \begin{subfigure}{\columnwidth} 
        \centering
        \includegraphics[width=0.95\textwidth]{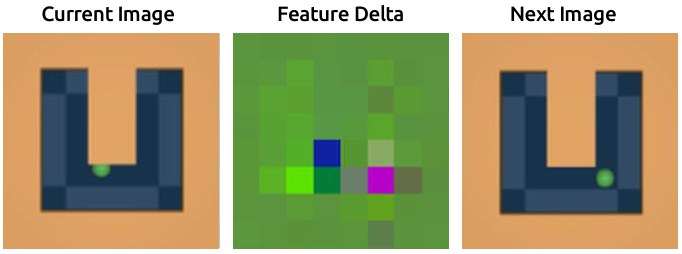}
        \caption{}
        \label{fig:intro_fig1_b}
    \end{subfigure}
    \vspace{-15pt}
    \caption{\textbf{(a) PCA visualization of internal feature evolution in a dense model. (b) PCA of the difference between consecutive ground-truth features.}
In (a), PCA projection of features from each predictor layer shows background regions remain largely static, revealing the computational redundancy of dense models. In (b), the PCA of the feature difference shows most regions (green) have near-zero change, demonstrating the inherent sparsity of physical dynamics.}
    \vspace{-15pt}
    \label{fig:intro_fig1}
\end{figure}



Endowing machines with the ability to perform autonomous planning and decision-making in complex, dynamic environments is a core objective of embodied intelligence research. In this pursuit, world models, which are capable of predicting future world states, play a pivotal role. By learning the dynamic laws of an environment directly from high-dimensional pixel inputs, world models enable an agent to mentally simulate the potential consequences of different action sequences without real-world interaction. This capability is the cornerstone of advanced planning algorithms such as Model Predictive Control (MPC) \citep{garcia1989model}, providing a powerful theoretical framework for solving complex robotic manipulation and navigation tasks \citep{liu2025aligning}.


\begin{figure}[t]
    \centering
    \includegraphics[width=0.85\columnwidth]{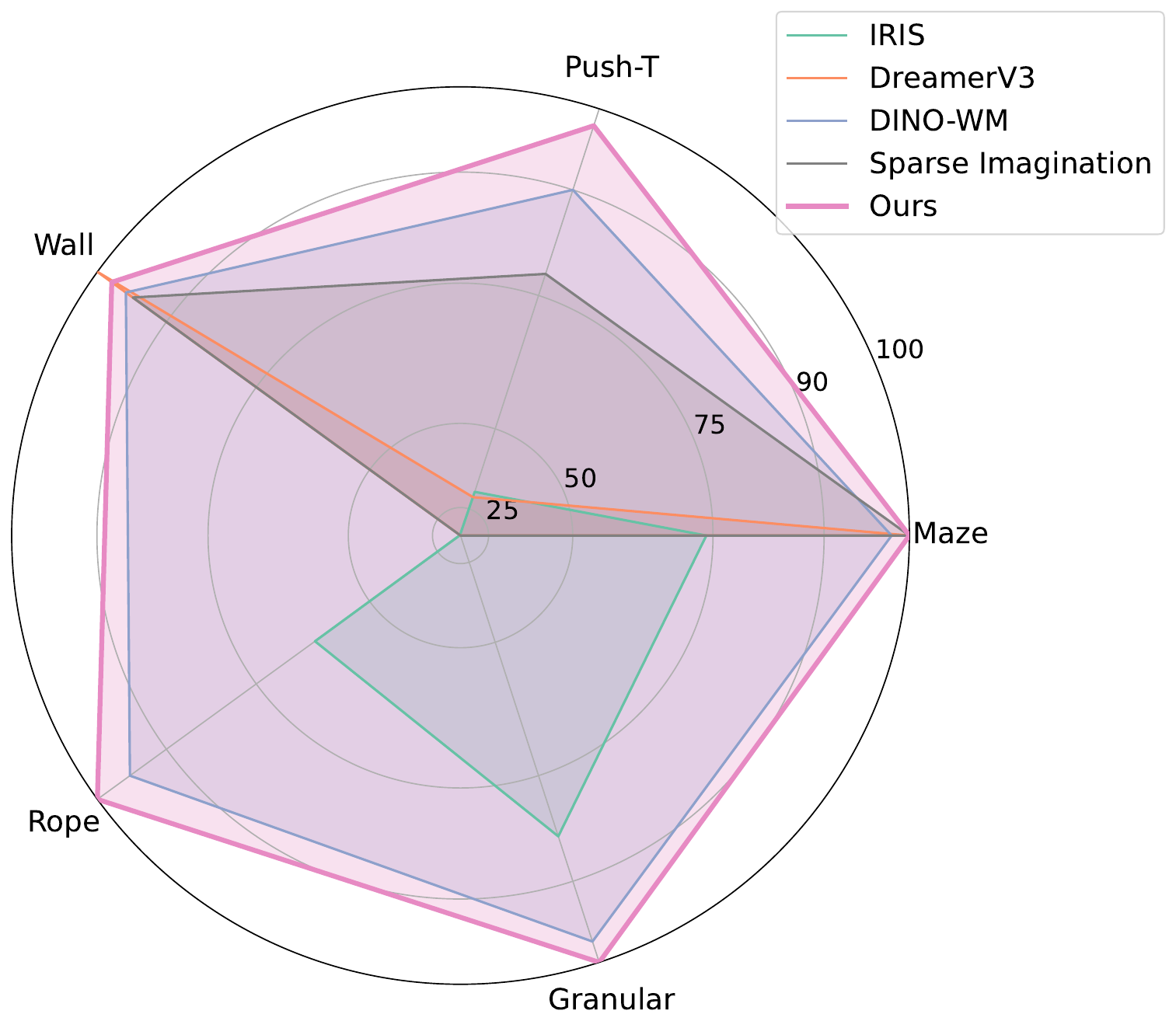}
    \caption{\textbf{Overall Performance on Key Benchmarks.} For comparability, Success Rates are plotted directly, while Chamfer Distance (CD) values are normalized via the formula $(\text{max}_{\text{CD}} - \text{CD}_{\text{i}}) / (\text{max}_{\text{CD}} - \text{min}_{\text{CD}}) \times 100$. }
    \label{fig:radar_summary}
    \vspace{-5pt}
\end{figure}

In recent years, building world models upon pre-trained visual representations (e.g., DINOv2 \citep{oquab2024dinov}) has become a frontier direction. Cutting-edge works like DINO-WM \citep{pmlr-v267-zhou25t} have demonstrated that architectures based on Vision Transformers (ViTs \citep{dosovitskiy2021an}) exhibit strong zero-shot planning capabilities by accurately capturing complex physical dynamics. However, this \textbf{indiscriminate} computational paradigm introduces a significant efficiency bottleneck: the model applies the same expensive self-attention computation to all image patches, regardless of whether they correspond to moving objects or static backgrounds. In most physical interaction scenarios, the regions undergoing actual change constitute only a small fraction, meaning the vast majority of computation is wasted on redundant recalculations for static backgrounds. For real-time MPC applications, which require hundreds or even thousands of simulations per second, prediction inference speed is critical for the deployment and application of advanced world models in real robotic systems. 
Indeed, our experiments reveal a significant practical bottleneck: even the current state-of-the-art dense model (DINO-WM) requires nearly two minutes (120 seconds) for a single MPC decision cycle on a representative manipulation task like Push-T. For many applications requiring continuous interaction with the physical world, such latency poses a fundamental challenge, providing direct motivation for our work.


To intuitively reveal the nature of this computational redundancy, we conducted an analysis of the internal workings of a SOTA dense model (DINO-WM) and the dynamic data it processes. As shown in Figure \ref{fig:intro_fig1_a}, we visualized the evolution of features in each layer of its predictor relative to the input using PCA. The results indicate that for the vast majority of background regions, the change in their features is almost negligible after passing through multiple layers of expensive self-attention computations. This confirms that dense models waste a significant amount of computational power on redundant recalculations for static regions. Furthermore, we find that the root cause of this computational waste lies in the inherent sparsity of physical dynamics. As depicted in Figure \ref{fig:intro_fig1_b}, we computed and visualized the difference between the feature maps of two consecutive frames. It was found that the features undergoing significant changes account for only a very small portion.


Based on this insight, we propose DDP-WM (Disentangled Dynamics Prediction World Model), an innovative framework that adheres to the design philosophy of allocating computational resources commensurate with the nature of the dynamics. This framework identifies the sparse regions where primary dynamics occur via a dynamic localization network and focuses computational resources on them using a powerful primary predictor. Concurrently, an efficient Low-Rank Correction Module (LRM) handles the context-driven background updates induced by the primary dynamics at a very low computational cost, thereby optimally allocating computational resources while providing a smooth optimization landscape for the planner. Figure~\ref{fig:radar_summary} provides a high-level summary of our method's comprehensive performance gains across key benchmarks.

Our main contributions can be summarized as follows:
\begin{itemize}
    \item We introduce the Disentangled Dynamics Prediction (DDP) paradigm, which posits that scene dynamics can be fundamentally decoupled into sparse, action-driven ``primary dynamics'' and broader ``context-driven background updates.''
    \item We propose DDP-WM, a novel architecture that instantiates the Disentangled Dynamics Paradigm (DDP). Its key innovation is the Low-Rank Correction Module (LRM), which leverages a unidirectional, causal cross-attention mechanism to efficiently capture background dynamics with minimal computational cost, thereby ensuring feature-space consistency.
    \item We demonstrate that DDP-WM establishes a new state of the art in both performance and efficiency. On the challenging Push-T benchmark, for instance, it improves the success rate from 90\% to a near-perfect 98\% while achieving an approximately 9x speedup. Our analysis further reveals that this closed-loop success is critically enabled by the smooth, tractable optimization landscape our method provides for the planner.
\end{itemize}

We will release our code, models, and supplementary materials to ensure the reproducibility of our results.

\section{Related Work}
\label{related_work}

\subsection{Visual World Models}
Model-based decision-making, where an agent makes informed choices by simulating the future, has been a long-standing goal in reinforcement learning and robotics \citep{NEURIPS2018_2de5d166, hafner2019planet}. The field has undergone a paradigm shift from early approaches that directly predicted pixels \citep{Kaiser2020Model} to modeling dynamics in a compact latent space. Starting with the pioneering work of \citet{NEURIPS2018_2de5d166}, PlaNet \citep{hafner2019planet} and the subsequent Dreamer series \citep{Hafner2020Dream, hafner2021mastering, hafner2025dreamerv3} have matured this latent dynamics modeling paradigm. They compress high-dimensional observations into a low-dimensional latent space using a Variational Autoencoder or similar methods, and then perform dynamics prediction with recurrent or sequence models. Subsequently, MWM \citep{seo2023masked} further demonstrated the effectiveness of decoupling representation and dynamics learning through masked autoencoders. However, these models rely on image reconstruction as a training objective, which can be affected by background noise in images, making it difficult to capture the fine-grained physical details required for high-precision robotic manipulation.

In recent years, Transformer-based sequence models, such as IRIS \citep{iris2023}, STORM \citep{zhang2023storm}, and V-JEPA 2 \citep{assran2025vjepa2}, have become the state-of-the-art approach due to their powerful modeling capabilities, achieving unprecedented high accuracy in capturing complex physical dynamics. An important trend is to perform dynamics prediction directly in the rich feature space provided by pre-trained visual models (e.g., DINOv2 \citep{oquab2024dinov}) to avoid information loss from reconstruction, as exemplified by \textbf{DINO-WM} \citep{pmlr-v267-zhou25t}. Meanwhile, large-scale video generation models, such as Genie \citep{pmlr-v235-bruce24a} and GAIA-2 \citep{russell2025gaia2controllablemultiviewgenerative}, are also considered a form of generalized world models. However, they focus more on open-ended video generation rather than the precise dynamics prediction for closed-loop planning that is the focus of this paper.

Despite the powerful performance of the aforementioned Transformer-based models, they invariably rely on dense self-attention computation over all visual tokens. Their computational complexity, which is quadratic with respect to the sequence length, severely hinders their application in real-time Model Predictive Control (MPC) that requires high-frequency simulations. Our work aims to significantly improve computational efficiency through a decoupled, sparse prediction framework, while maintaining or even improving upon their performance.

A broader discussion of related work, including efficiency optimizations for general Transformers and other sparse or structured modeling approaches, is provided in Appendix~\ref{related_work}.

\begin{figure*}[t]
    \centering
    \includegraphics[width=1\textwidth]{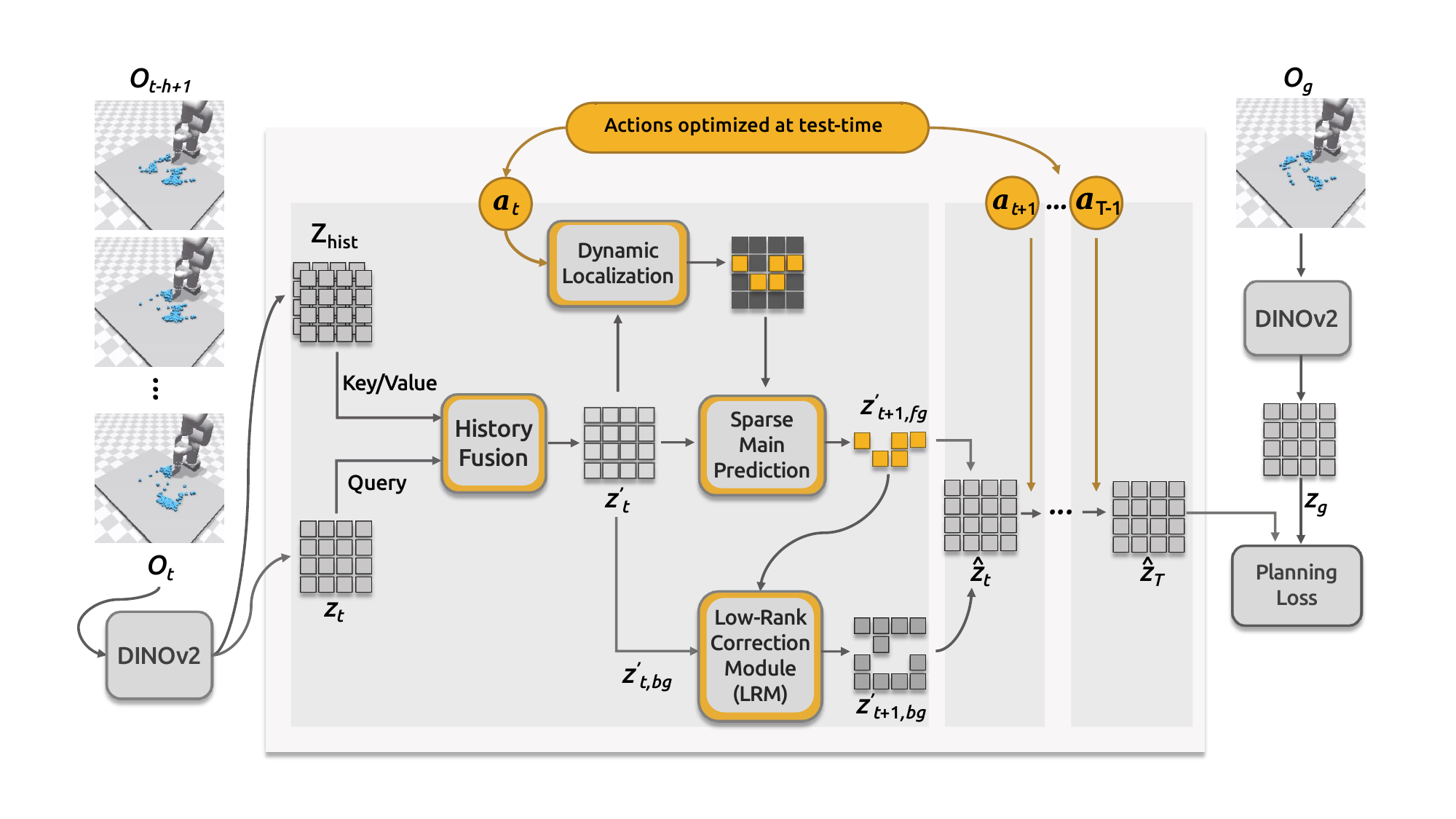}
\vspace{-30pt}\caption{\textbf{Overview of the DDP-WM Framework.} Our framework performs prediction through a four-stage decoupled process. \textbf{(1) Historical Information Fusion:} The features of the current frame, $\mathbf{z}_t$, query historical frame features $\mathbf{Z}_{\text{hist}}$ via cross-attention to obtain temporally-aware augmented features $\mathbf{z}'_t$. \textbf{(2) Dynamic Localization:} A lightweight network receives $\mathbf{z}'_t$ and the action $\mathbf{a}_t$ to predict a sparse mask $\mathbf{M}$ that contains only the primary dynamic regions. \textbf{(3) Sparse Primary Dynamics Prediction:} A powerful primary predictor focuses all computation on the sparse foreground features identified by $\mathbf{M}$ to predict the next frame's foreground features, $\mathbf{z}'_{t+1, \text{fg}}$, with high precision. \textbf{(4) Contextual Background Update:} The background features of the current frame, $\mathbf{z}_{t, \text{bg}}$, are updated at a very low cost by querying the newly predicted foreground features $\mathbf{z}'_{t+1, \text{fg}}$ via cross-attention. Finally, the updated foreground and background are combined to constitute the complete latent state of the next frame.}
\vspace{-5pt}
    \label{fig:fig2}
\end{figure*}

\section{Methodology}

To address the inherent contradiction between computational efficiency and planning performance in existing world models, we propose an innovative decoupled dynamics prediction framework. The core idea is to decompose complex visual dynamics into two sub-problems of different natures and to design specialized, efficient computational modules for them. Figure \ref{fig:fig2} provides a complete overview of the decoupled dynamics prediction framework.

\subsection{Background and Problem Formulation}

We formalize the visual control task as a Partially Observable Markov Decision Process (POMDP). Within this framework, a world model built upon pre-trained features typically consists of two core components:

\begin{enumerate}
    \item A fixed, pre-trained observation model $g_\phi$: It is responsible for mapping high-dimensional image observations $o_t \in \mathbb{R}^{H \times W \times 3}$ to a series of latent features (patch tokens) containing rich spatial information, i.e., $z_t = g_\phi(o_t)$, where $z_t \in \mathbb{R}^{N \times D}$. In this paper, we adopt a frozen DINOv2 as our observation model, which is consistent with the choice in cutting-edge works like DINO-WM.
    \item A learnable transition model $f_\theta$: Its task is to predict the latent state at the next time step, $\hat{z}_{t+1} = f_\theta(z_{\le t}, a_{\le t})$, based on the history of state-action sequences $(z_{\le t}, a_{\le t})$.
\end{enumerate}

Our core contribution revolves around designing a novel transition model, $f_\theta$, that far surpasses existing state-of-the-art (SOTA) methods in both performance and efficiency.

\subsection{Core Insight: Decoupling Primary Dynamics and Context-driven Background Updates}

Dense world models, such as DINO-WM, predict the future by applying a uniform transformation to all tokens, which leads to massive computational redundancy in scenarios with sparse changes. More severely, this fully-connected computation pattern can cause the model to learn non-causal spurious correlations within the scene, harming its generalization ability and robustness. An intuitive optimization is to employ sparse computation, i.e., only predicting for regions undergoing change while performing feature copying for static regions. However, our preliminary experiments (detailed in Section ~\ref{ablation}) reveal a critical phenomenon: while such simple sparse models achieve lower error in open-loop prediction, their planning success rate in closed-loop Model Predictive Control (MPC) plummets.

We argue that a key factor in this problem is an overlooked phenomenon: in pre-trained representations based on self-attention mechanisms (like DINOv2), any local feature implicitly encodes its relationship with the global context. Consequently, when a primary object moves, even if the pixels in static regions remain unchanged, their features must undergo subtle adjustments due to the change in spatial context. We term this context-aware adjustment of the background, triggered by foreground changes, as context-driven background updates.

Simple sparse models, with their copy-paste rule, violate this intrinsic property of the feature space. This results in discontinuous cliffs in the cost landscape provided to the planner, which is the root cause of planning failures.


Based on this insight, we decouple and separately model two types of dynamics:
\begin{enumerate}
    \item \textbf{Primary Dynamics:} High-frequency, non-linear changes on foreground objects caused by direct physical interactions, reflecting the core causal chain of the scene.
    \item \textbf{Context-driven Background Updates:} Low-frequency, context-driven feature adjustments in background regions, induced by the primary dynamics. We further make a key assumption that this seemingly complex global adjustment is inherently \textbf{low-rank}. More formally, this posits that the set of all background update vectors, $\{\Delta\mathbf{z}_i\}$, lies in a low-dimensional subspace, which is mathematically equivalent to their corresponding Gram matrix being of low rank. We provide decisive empirical evidence for this foundational assumption in Appendix~\ref{sec:low_rank_analysis}, where a Principal Component Analysis (PCA) directly inspects the effective rank of this structure.

\end{enumerate}

To this end, we have designed a decoupled dynamics prediction framework.

\subsection{Decoupled Dynamics Prediction Framework}

Our framework (as shown in Figure \ref{fig:fig2}) achieves efficient and high-fidelity dynamics prediction through a four-stage process. It first injects temporal dynamics into the current state through a historical information fusion module, followed by the Dynamic Localization Network, Sparse Primary Dynamics Predictor, and Low-Rank Correction Module, which jointly complete the prediction for the next frame.

\subsubsection{Stage 1: Historical Information Fusion Module}

To enable the model to understand higher-order dynamics such as velocity and acceleration, we introduce an efficient historical information fusion module before the core prediction process begins.

\begin{itemize}
    \item \textbf{Mechanism and Efficiency:} Unlike dense models such as DINO-WM, which simply stack the features of all historical frames and feed them into a full Transformer, our method accomplishes the injection of historical information via a single layer of cross-attention (CA).
        \begin{itemize}
            \item \textbf{Query:} The latent features of the current frame, $\mathbf{z}_t$.
            \item \textbf{Key/Value:} The set of latent features from all historical frames, $\mathbf{Z}_{\text{hist}} = \{\mathbf{z}_{t-h+1}, ..., \mathbf{z}_{t-1}\}$.
        \end{itemize}
    \item \textbf{Workflow:} Each feature vector of the current frame queries the complete history information pool via cross-attention to aggregate relevant temporal dynamics, and updates itself through a residual connection:
    \begin{equation}
    \mathbf{z}'_t = \mathbf{z}_t + \text{CA}(\text{Q}=\mathbf{z}_t, \text{K}=\mathbf{Z}_{\text{hist}}, \text{V}=\mathbf{Z}_{\text{hist}})
    \end{equation}\textbf{}
    The features $\mathbf{z}'_t$, augmented by this module, contain an implicit encoding of the current dynamics and serve as the input for the subsequent prediction stages.
\end{itemize}

\subsubsection{Stage 2: Dynamic Localization Network}

This module is responsible for efficiently and accurately identifying the sparse regions where primary dynamics will occur in the next frame.

\begin{itemize}
    \item \textbf{Architecture and Task:} This module is responsible for efficiently and accurately identifying the sparse regions where primary dynamics will occur. A dedicated ViT receives the temporally-aware current state latent features $\mathbf{z}'_t$ and the action $\mathbf{a}_t$. To improve localization accuracy, it predicts change probabilities for the corresponding $2 \times 2$ sub-regions of each image patch, outputting a probability map $P_{\text{sub}} \in \mathbb{R}^{N \times 4}$. The final sparse binary mask $\mathbf{M} \in \{0, 1\}^N$ is then generated by thresholding these probabilities: a patch is marked as changed if any of its sub-regions' predicted change exceeds a preset threshold $\tau$. This process is defined as:
    \begin{equation}
    \label{eq:mask_generation}
    m_i = 
    \begin{cases} 
    1 & \text{if } \max(P_{\text{sub}, i}) > \tau \\
    0 & \text{otherwise}
    \end{cases}
    \quad \text{for } i=1, \dots, N
    \end{equation}
    where $m_i$ is the $i$-th element of the mask $\mathbf{M}$, and $P_{\text{sub}, i}$ denotes the 4 change probabilities for the sub-regions of the $i$-th patch.
\end{itemize}

\subsubsection{Stage 3: Sparse Primary Dynamics Predictor}
\label{subsubsec:SPDP}
This module is the primary computational unit, responsible for modeling the primary dynamics identified by the mask $\mathbf{M}$ with high precision.


\textbf{Implementation:} From the complete, history-fused input features $\mathbf{z}'_t$, we use the mask $\mathbf{M}$ to extract the subset of dynamic patch tokens, $\mathbf{z}'_{t, \text{fg}}$. A powerful primary predictor (e.g., a ViT-based model) then focuses its computation entirely on this sparse subset to predict the high-precision next-frame foreground features $\mathbf{z}'_{t+1, \text{fg}}$.

\textbf{Adaptive Sparse Size:} We employ an adaptive sizing strategy to handle dynamically varying sparse inputs, balancing hardware efficiency with computational accuracy. The detailed mechanism is deferred to Appendix~\ref{app:adaptive_size}.

\subsubsection{Stage 4: Low-Rank Correction Module (LRM)}


This module is designed to efficiently model the context-driven background updates induced by the primary dynamics, at a very low computational cost. Its core design architecturally embodies an inductive bias that mimics the unidirectional causal flow of physics: the primary dynamics are computed first, and the background update must proceed based on their result, rendering the information flow irreversible.

Specifically, we employ a single-layer cross-attention mechanism, where the information flow is asymmetrically designed as follows:
\begin{itemize}
    \item \textbf{Query:} The set of background patch features $\mathbf{z}'_{t, \text{bg}}$.
    \item \textbf{Key/Value:} The newly predicted foreground patch features $\mathbf{z}'_{t+1, \text{fg}}$.
\end{itemize}
This asymmetric setup forces each background token to passively query the completed foreground predictions, thereby modeling the background update as a direct consequence of the primary dynamics. Each background patch aggregates information about the foreground changes via the attention mechanism to generate an update vector. Finally, the new background features are updated through a residual connection:
\begin{equation}
\mathbf{z}'_{t+1, \text{bg}} = \mathbf{z}'_{t, \text{bg}} + \text{CA}(\text{Q}=\mathbf{z}'_{t, \text{bg}}, \text{K}=\mathbf{z}'_{t+1, \text{fg}}, \text{V}=\mathbf{z}'_{t+1, \text{fg}})
\end{equation}
This operation achieves a self-consistent update of the background features at a very low computational cost, providing a smooth optimization landscape for the downstream planner.


\begin{table*}[]
    \centering
    \caption{Comparison of MPC planning performance in the five simulated environments.}
    \label{tab:perf_comp}
    \begin{tabular}{lccccc}
        \toprule
        Model & PointMaze (SR $\uparrow$) & Push-T (SR $\uparrow$) & Wall (SR $\uparrow$) & Rope (CD $\downarrow$) & Granular (CD $\downarrow$) \\
        \midrule
        IRIS & 74\% & 32\% & 4\% & 1.11 & 0.37 \\
        DreamerV3 & 100\% & 30\% & 100\% & 2.49 & 1.05 \\
        Sparse Imagination & 100\% & 78.3\% & 95\% & - &- \\
        DINO-WM & 98\% & 90\% & 96\% & 0.41 & 0.26 \\
        \textbf{DDP-WM (Ours)} & \textbf{100\%} & \textbf{98\%} & \textbf{98\%} & \textbf{0.31} & \textbf{0.24} \\
        \bottomrule
        \bottomrule
    \end{tabular}
            \vspace{-5pt}
\end{table*}

\begin{figure}[t]
    \centering
    \includegraphics[width=0.95\columnwidth]{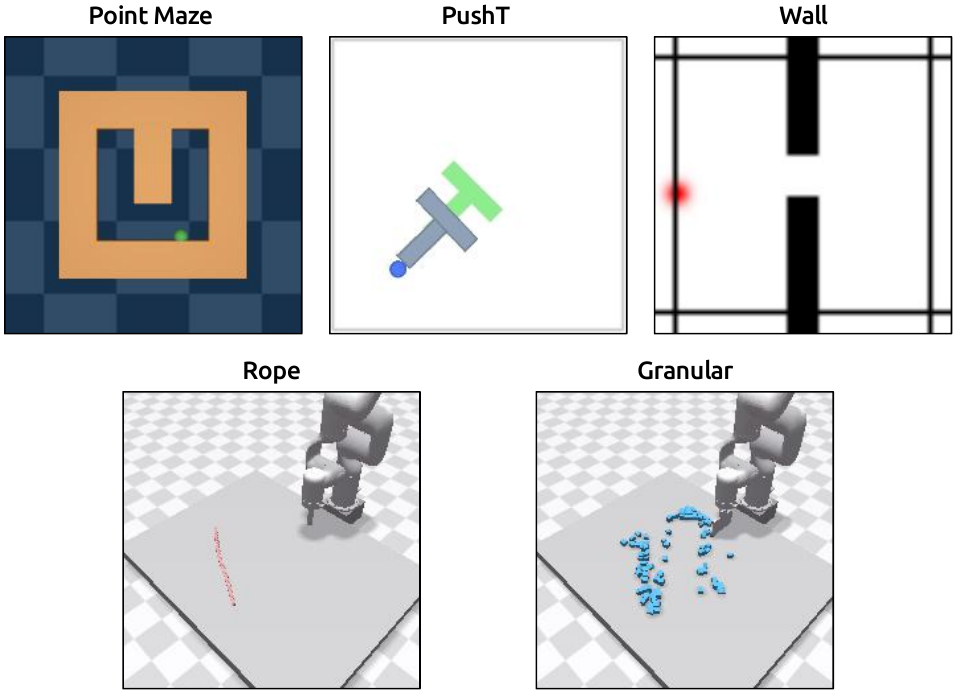} 
    \caption{\textbf{Overview of Evaluation Environments.} Sample frames from the five simulated task domains used in our experiments.}
    \label{fig:env_overview}
    \vspace{-5pt}
\end{figure}

\subsection{Planning with DDP-WM: Model Predictive Control}
\label{subsec:mpc}

We embed the trained DDP-WM as the dynamics model within a Model Predictive Control (MPC) framework for online trajectory planning. We employ the standard Cross-Entropy Method (CEM) as the optimizer. The detailed algorithm is deferred to Appendix~\ref{app:cem}.

\textbf{Sparse MPC Cost Mask.} For the cost function design, the conventional approach is to compute the Mean Squared Error (MSE) between all features of the predicted final state and the goal state. However, we find that this dense computation introduces unnecessary noise. Therefore, we propose a \textbf{Sparse MPC Cost Mask} strategy. Specifically, we first compute the pixel-wise difference between the current observation image and the goal image to generate a binary mask $\mathbf{M}_{\text{task}}$, which has a value of 1 only in the differing regions. When calculating the MPC cost, we only consider the feature error in these task-relevant regions:
\begin{equation}
\mathcal{L}_{\text{MPC}} = \text{MSE}(\hat{\mathbf{z}}_{T} \odot \mathbf{M}_{\text{task}}, \mathbf{z}_{\text{goal}} \odot \mathbf{M}_{\text{task}})
\end{equation}
This strategy focuses the planner's cost evaluation on the regions that truly need to change, filtering out interference from the static background and making the optimization process more efficient and stable. In the ablation studies in Section ~\ref{experiments}, we will quantitatively analyze the effectiveness of this strategy.

\section{Experiments}
\label{experiments}

\subsection{Experimental Setup}
\textbf{Environments.} To comprehensively evaluate the generalization ability of our method, we select five simulated environments with diverse dynamic characteristics and task complexities (shown in Figure~\ref{fig:env_overview}), covering a wide range of scenarios from simple navigation to complex physical interactions:
\begin{itemize}
    \item \textbf{PointMaze \& Wall:} Two 2D navigation tasks to evaluate the model's capabilities in basic kinematics and spatial reasoning.
    \item \textbf{Push-T:} A representative table-top manipulation task that requires the model to understand rigid-body contact dynamics to achieve precise pose control.
    \item \textbf{Rope \& Granular:} Two more challenging manipulation tasks involving the complex dynamics of a deformable body (rope) and a multi-body system (granular materials), respectively, placing higher demands on the model's understanding of physics.
\end{itemize}

\textbf{Evaluation Metrics.} We primarily focus on metrics across the following three dimensions:
\begin{itemize}
    \item \textbf{Model Predictive Control Success Rate (SR $\uparrow$):} For the navigation and Push-T tasks, we evaluate the agent's success rate in reaching the designated goal state under MPC planning.
    \item \textbf{Chamfer Distance (CD $\downarrow$):} In tasks such as Rope and Granular, where defining a binary success state is difficult, we evaluate the Chamfer Distance between the final state and the goal state. A lower value for this metric signifies more precise manipulation.
    \item \textbf{Computational Efficiency:} We measure efficiency gains in terms of theoretical Floating-Point Operations (FLOPs), single-step inference throughput (Throughput), and single MPC decision time (Latency).
\end{itemize}

\textbf{Baselines.} Our primary baseline is DINO-WM, the current state-of-the-art dense world model based on pre-trained features. To ensure a fair and direct comparison, all our experimental environments, datasets, and the core parameters of the MPC planner (CEM) strictly adhere to the official DINO-WM settings. This allows us to precisely attribute the differences in performance and efficiency to the fundamental distinction between our proposed decoupled dynamics framework and the dense framework.

\textbf{Implementation Details.} Specific implementation details, including model architectures, training hyperparameters, and planner parameters, are deferred to Appendix~\ref{app:hyperparams}.

\subsection{Quantitative Comparison: Planning Performance and Computational Efficiency}

\subsubsection{Planning Performance}

We conducted a comprehensive performance comparison between our full method (Ours) and the DINO-WM across the five environments. The results are presented in Table~\ref{tab:perf_comp}.

Table~\ref{tab:perf_comp} shows that our method matches or surpasses the SOTA dense model on all tasks. The most significant gain is on the challenging Push-T task, where our 98\% success rate substantially outperforms DINO-WM's 90\%, demonstrating that our approach provides the planner with higher-quality future predictions.

\subsubsection{Computational Efficiency}

\begin{table}[t]
    \centering
    \caption{Comparison of theoretical computational cost (FLOPs) for a single forward inference step.}
    \vspace{-5pt}
    \label{tab:flops}
    \begin{tabular}{lccc}
        \toprule
        Task & DINO-WM & Ours & FLOPs Reduction \\
        \midrule
        Push-T & 23 G & \textbf{2.5 G} & \textbf{9.2$\times$} \\
        Wall & 7.8 G & \textbf{2.5 G} & \textbf{3.1$\times$} \\
        \bottomrule
    \end{tabular}
            \vspace{-5pt}
\end{table}

\begin{table}[t]
    \centering
    \caption{Comparison of single-step inference throughput (samples/sec, $\uparrow$). Tested on a single NVIDIA 2080 Ti with a batch size of 128.}
        \vspace{-5pt}
    \label{tab:throughput}
    \begin{tabular}{lccc}
        \toprule
        Task & DINO-WM & Ours & Speedup \\
        \midrule
        Push-T & 170 & \textbf{1563} & \textbf{9.2$\times$} \\
        Wall & 802 & \textbf{2170} & \textbf{2.7$\times$} \\
        \bottomrule
    \end{tabular}
            \vspace{-5pt}
\end{table}

As shown in Table~\ref{tab:flops}, our method achieves a massive reduction in theoretical computational cost (FLOPs). In the dynamically complex Push-T task, DDP-WM's computational cost is only one-tenth that of the dense model.

This theoretical advantage translates directly into faster practical inference speed. In the single-step inference throughput test (Table~\ref{tab:throughput}), DDP-WM achieves a 9.2x improvement on the Push-T task.

This efficiency advantage is further reflected in the complete MPC decision loop (Table~\ref{tab:latency}). For the Push-T task, which requires 30 CEM iterations, our decision time is only 16 s, a 7.5x speedup compared to DINO-WM's 120 s, enabling higher-frequency control.

\begin{table}[t]
    \centering
    \caption{Comparison of single MPC decision loop time (s, $\downarrow$). Tested on a single NVIDIA A5880. }
        \vspace{-5pt}
    \label{tab:latency}
    \begin{tabular}{lccc}
        \toprule
        Task / Iterations & DINO-WM & Ours & Speedup \\
        \midrule
        PointMaze / 10 & 39 s & \textbf{5.5 s} & \textbf{7.1$\times$} \\
        Push-T / 30 & 120 s & \textbf{16 s} & \textbf{7.5$\times$} \\
        Wall / 10 & 12 s & \textbf{4.2 s} & \textbf{2.9$\times$} \\
        Rope / 10 & 12 s & \textbf{4.3 s} & \textbf{2.8$\times$} \\
        Granular / 30 & 35 s & \textbf{13 s} & \textbf{2.7$\times$} \\
        \bottomrule
    \end{tabular}
            \vspace{-5pt}
\end{table}

These data demonstrate that our decoupled framework leads to significant gains in computational efficiency while also improving planning performance.

\begin{table}[t]
    \centering
    \caption{Impact of localization quality on 5-step open-loop prediction pixel error (see Appendix~\ref{app:pixel_error} for details) in Push-T, $\downarrow$.}
\vspace{-5pt}\label{tab:ablation_pixel_error}
    \begin{tabular}{lcc}
        \toprule
        Localization Method & w/o LRM & w/ LRM \\
        \midrule
        Patch-level & 788 & 468 \\
        High-Precision & 427 & 361 \\
        \bottomrule
    \end{tabular}
        \vspace{-5pt}
\end{table}

\begin{table}[t]
    \centering
    \caption{Comparison of 5-step open-loop prediction pixel error (number of error pixels, $\downarrow$).}
    \label{tab:openloop}
    \begin{tabular}{lccc}
        \toprule
        Model & PointMaze & Push-T & Wall \\
        \midrule
        DINO-WM (Dense) & 81 & 524 & 111 \\
        DDP-WM (Ours) & 36 & 361 & 9 \\
        Naive Sparse (w/o LRM) & 41 & 427 & 15 \\
        \bottomrule
    \end{tabular}
\end{table}

\begin{table}[t]
    \centering
    \caption{Ablation study results on the Push-T task.}
    \label{tab:ablation}
    \begin{tabular}{lccc}
        \toprule
        Model Variant & LRM (c) & MPC Mask (m) & SR $\uparrow$ \\
        \midrule
        DDP-WM (Ours) & $\checkmark$ & $\checkmark$ & \textbf{98\%} \\
        & $\checkmark$ & $\times$ & 90\% \\
        & $\times$ & $\checkmark$ & 70\% \\
        Naive Sparse & $\times$ & $\times$ & 62\% \\
        \bottomrule
    \end{tabular}
\end{table}

\begin{table}[t]
    \centering
    \caption{Ablation study on the High-Precision Localization mechanism on the Push-T task (Mask Quality).}
\label{tab:ablation_localization}
    \begin{tabular}{lccc}
        \toprule
        Localization Method & IoU $\uparrow$ & Precision $\uparrow$ & Recall $\uparrow$ \\
        \midrule
        Patch-level & 0.3446 & 0.6971 & 0.5172 \\
        High-Precision & 0.8935 & 0.9058 & 0.9845 \\
        \bottomrule
    \end{tabular}
\end{table}

\begin{figure}[t]
    \centering
    \begin{subfigure}[b]{0.22\textwidth}
        \centering
        \includegraphics[width=1\textwidth]{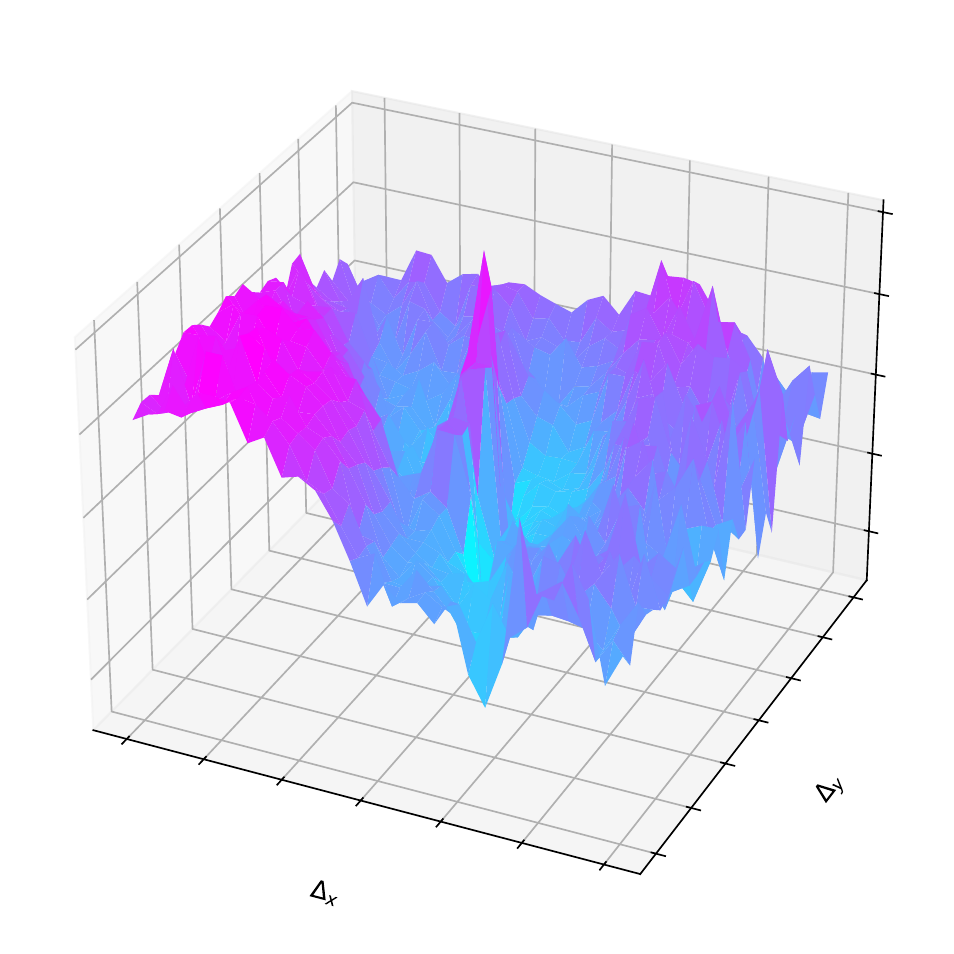}
        \caption{} 
    \end{subfigure}
    \begin{subfigure}[b]{0.22\textwidth}
        \centering
        \includegraphics[width=\textwidth]{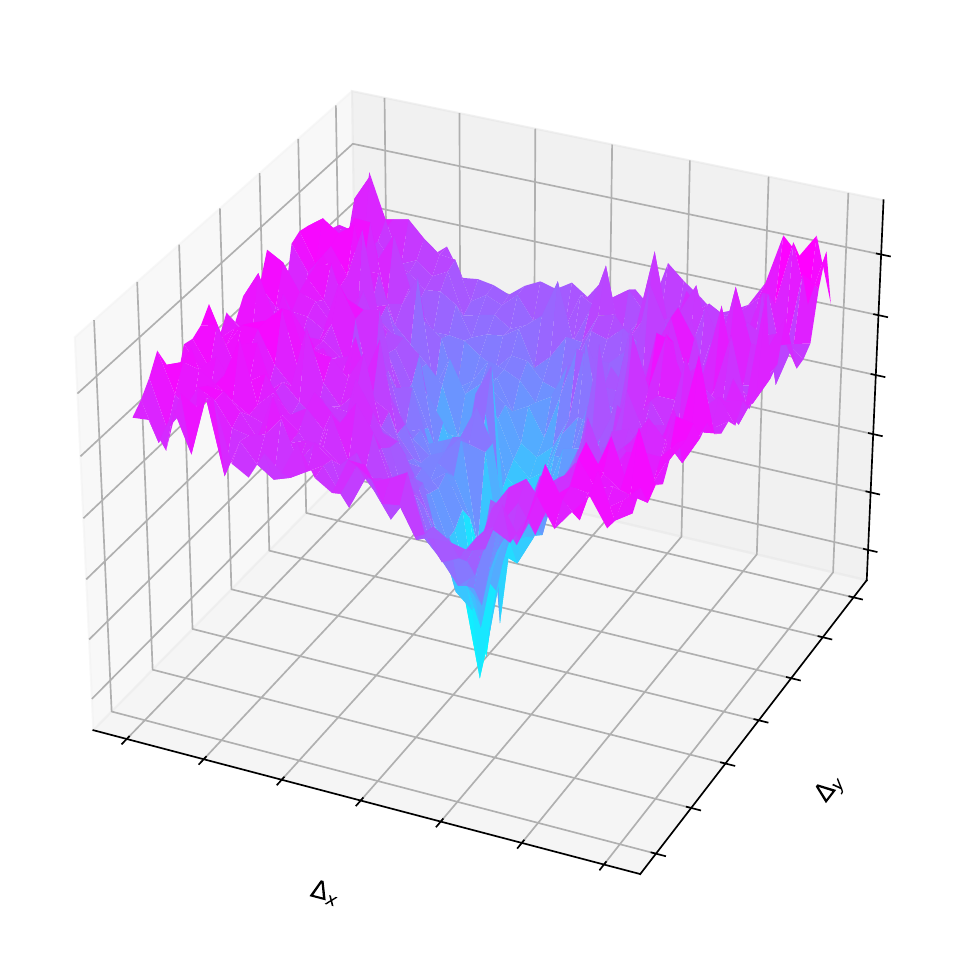}
        \caption{} 
    \end{subfigure}
   \caption{\textbf{Comparison of MPC cost function landscapes created by different models on the Push-T task.} Both plots show the cost surfaces after 2D perturbation of the action space. \textbf{(Left) Naive Sparse model (w/o LRM):} The cost landscape is rugged and noisy, trapping the optimizer in local minima. \textbf{(Right) Our DDP-WM model (w/ LRM):} In contrast, the landscape is smooth with a clear, funnel-shaped global minimum, enabling efficient optimization.}
    \label{fig:cost_landscapes}
\end{figure}

\subsection{Ablation Study}
\label{ablation}

To systematically validate the necessity of each design within our framework, we conducted a series of detailed ablation studies on the most challenging Push-T task. We primarily investigate the effectiveness of two core components: the LRM, the Sparse MPC Cost Mask (MPC Mask) and the Dynamic Localization Network.

We compare our full method (denoted as $\checkmark$ for both components) with variants where these components are removed. The results are presented in Table~\ref{tab:ablation}.

We further ablate the high-precision localization mechanism. As shown in Table~\ref{tab:ablation_localization}, the mechanism brings a decisive improvement to the quality of mask prediction. More importantly, this improvement in localization accuracy translates directly into more accurate overall dynamics prediction, as shown in Table~\ref{tab:ablation_pixel_error}.

\subsection{Analysis}

The ablation study reveals a core question: Why does the LRM bring such a significant improvement in closed-loop performance? To explain this at a deeper level, we design experiments to probe and visualize the optimization landscape that different models create for the planner.


\subsubsection{The Paradox of Open-Loop Prediction}
To investigate this, we first compare the open-loop prediction accuracy. For this specific analysis, we intentionally measure error in pixel space (see Appendix~???\ref{app:pixel_error} for calculation details). This choice is crucial for a fair ablation, as the `Naive Sparse` model does not predict background tokens, and a direct feature MSE comparison would be inequitable. Pixel error, in contrast, provides an objective measure of physical accuracy. We task the models with predicting a 5-step trajectory; the results are in Table~???\ref{tab:openloop}.

As shown in Table~\ref{tab:openloop}, a noteworthy discrepancy emerges: in open-loop prediction, the error of the Naive Sparse model (w/o LRM) is nearly identical to that of our full DDP-WM model, and both are significantly lower than the dense DINO-WM. This indicates that the LRM provides almost no help in improving open-loop prediction accuracy. Why, then, is this module, seemingly useless in open-loop, the key to success in closed-loop planning?

\subsubsection{Optimization Landscape Smoothness}

We posit that the stark contrast between open-loop and closed-loop performance lies in the smoothness of the optimization landscape that different models create for the planner. To verify this hypothesis, we designed a series of experiments to visualize this cost function landscape. We first performed one-dimensional action space scan experiments, where we applied perturbations to a single dimension of a successful action sequence and observed the change in cost. To obtain a more comprehensive view, we further conducted a two-dimensional scan: we fixed the first 4 steps of a 5-step action sequence and applied grid-sampled perturbations to the last action along two orthogonal dimensions, plotting the MPC cost corresponding to each perturbed action as a 3D surface map.


As shown in Figure~\ref{fig:cost_landscapes}, the results provide clear and intuitive evidence. The landscape generated by the Naive Sparse model (left) is highly rugged and noisy, lacking a stable global minimum for convergence. On such a landscape, any sampling-based optimizer is akin to blindly searching in a treacherous, trap-filled terrain, making it extremely easy to fall into local optima. This provides a clear explanation for its failure in closed-loop planning.

In stark contrast, the cost landscape generated by our full DDP-WM model (right) is exceptionally smooth and exhibits a clear, funnel-like macroscopic structure, with a single, deep global minimum at its center. This landscape provides a clear ``gravity well" for the optimizer, enabling it to converge stably and efficiently towards the optimal solution.

%
\section{Conclusion}

The core contribution of this paper is an efficient world model paradigm that focuses computation on sparse primary dynamics via localization. We identify that the key to making such a sparse approach succeed in closed-loop planning is to solve the un-smooth optimization landscape problem it introduces. Our proposed Low-Rank Correction Module (LRM) is the architectural solution designed for this purpose, ensuring a plannable landscape by maintaining feature-space consistency. This synergy between sparse prediction and landscape-smoothing correction is responsible for DDP-WM's state-of-the-art results in both performance and speed. Our DDP-WM establishes a promising path for
developing efficient, high-fidelity world models.

\bibliography{paper}
\bibliographystyle{icml2026}

\newpage
\appendix
\onecolumn
\newpage
\appendix
\onecolumn

\newpage
\appendix
\onecolumn

\section{Empirical Verification: The Low-Rank Structure of Background Updates}
\label{sec:low_rank_analysis}

Our method's core hypothesis is that the context-driven background updates, induced by primary dynamics, possess an inherently low-rank structure. To both (1) empirically validate this central assumption and (2) examine whether our LRM has successfully learned this structure, we conducted a parallel Principal Component Analysis (PCA) on the ground-truth background updates and the updates generated by our LRM.

\begin{figure}[h]
    \centering
    \includegraphics[width=0.9\columnwidth]{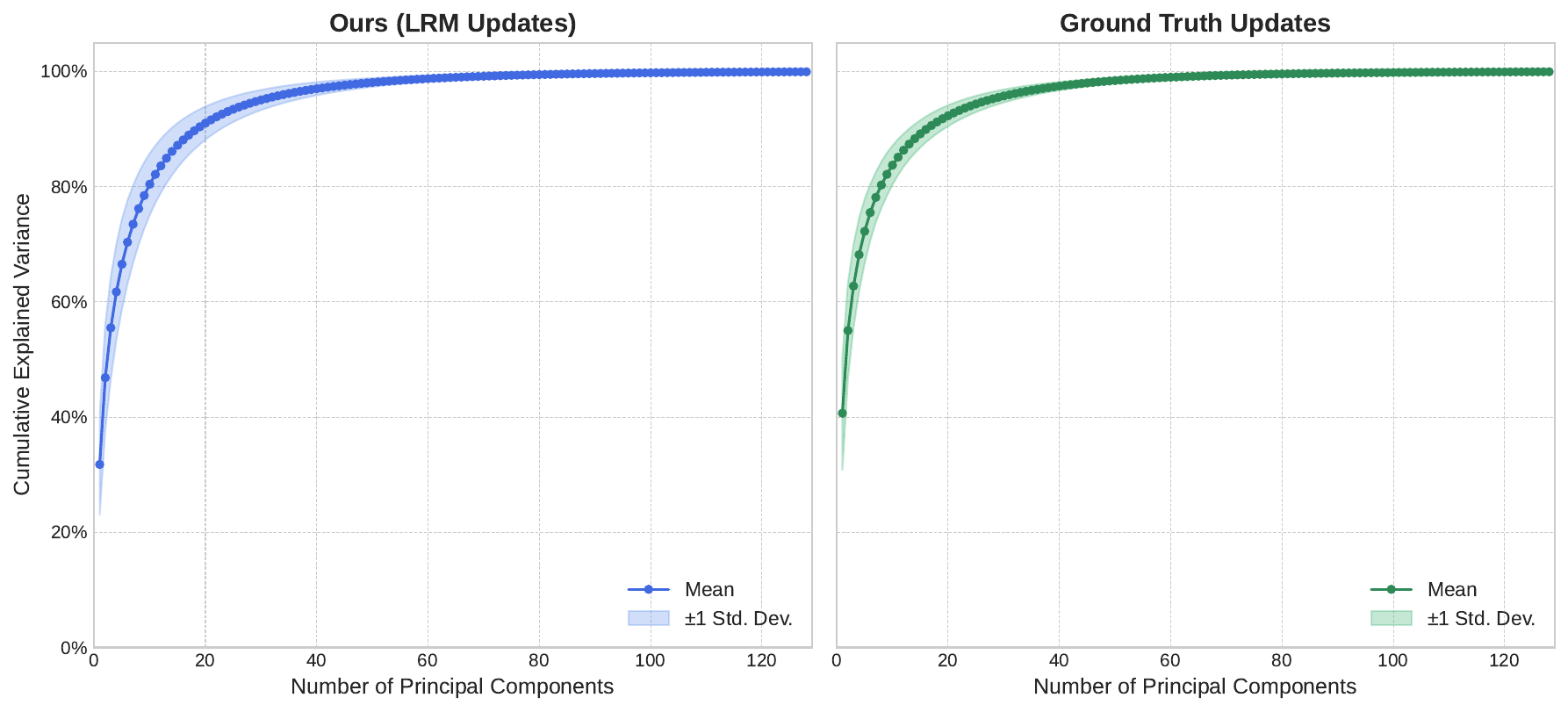}
    \caption{\textbf{LRM successfully learns the true low-dimensional structure.} Parallel PCA performed on (Left) the background updates predicted by our LRM and (Right) the ground-truth background updates, on the Push-T task. The remarkable similarity between the two curves indicates that our model has accurately captured and replicated the inherent low-rank nature of the physical dynamics.}
    \label{fig:pca_variance_comparison}
    \vspace{-15pt}
\end{figure}

As depicted in Figure~\ref{fig:pca_variance_comparison}, the results provide decisive, twofold evidence for our claims.
First, the cumulative explained variance curve for the Ground Truth Updates (right plot) exhibits a sharp rise followed by rapid saturation, strongly verifying our core hypothesis that the true physical dynamics possess a very low intrinsic dimensionality.
Second, and most critically, the PCA curve for the updates generated by our LRM (left plot) is strikingly similar to that of the ground truth.

This high degree of alignment convincingly demonstrates that our LRM has successfully and faithfully learned to replicate the intrinsic low-dimensional structure present in the true world dynamics. This not only provides the strongest justification for naming our module the "Low-Rank Correction Module" but also fundamentally explains our method's effectiveness and efficiency: it precisely exploits the inherent "simplicity" underlying the physical dynamics.

\section{Additional Related Work}
\label{related_work}

\subsection{Efficiency Optimization of Transformer Models}

With the popularization of Vision Transformers (ViT) \citep{dosovitskiy2021an}, their high computational cost has spurred a large body of research on efficiency optimization. One class of methods involves general token sparsification techniques, which dynamically remove tokens deemed unimportant during inference. These methods either learn token importance through a lightweight network \citep{rao2021dynamicvit, NEURIPS2021_6a30e32e}, or perform pruning based on heuristic rules such as attention scores \citep{wang2020glance}. Another class of methods introduces masked autoencoding mechanisms during the training phase, such as MAE \citep{he2022masked} and VideoMAE \citep{tong2022videomae}, which have shown that models can recover full information from partially visible tokens.

These ideas have also been adapted to world models to improve efficiency. For example, MaskViT \citep{gupta2023maskvit} applies the masked generation paradigm to video prediction and accelerates generation through iterative decoding. Closer to our work is Sparse Imagination \citep{chun2025sparseimaginationefficientvisual}, which randomly drops a portion of image patch tokens during the imagination (rollout) phase of MPC to accelerate computation. Another related work \citep{micheli2024efficient} shortens the sequence length by predicting the delta between consecutive states.

However, these general or random sparsification methods do not fully leverage the intrinsic structure of physical dynamics in robotic interaction scenarios. Unlike them, our DDP-WM proposes a structured sparsity scheme grounded in physical insight, designed to leverage the intrinsic structure of physical dynamics in robotic interaction. The framework explicitly decouples the dynamics into primary dynamics and context-driven background updates, and assigns specialized computational modules to each.

\subsection{Sparse and Structured Dynamics Modeling}

Leveraging the intrinsic sparsity of dynamics to build more efficient and interpretable models is a natural and attractive research direction. Among these, a mainstream paradigm is object-centric learning. Starting from the pioneering Slot Attention \citep{locatello2020object}, many works have attempted to decompose the world into independent objects and model dynamics at the object level, such as C-SWMs \citep{Kipf2020Contrastive}, FOCUS \citep{ferraro2025focus}, and OC-STORM \citep{zhang2025objectsmatterobjectcentricworld}. These methods, by introducing inductive biases, hold the promise of enhancing the model's generalization capabilities and data efficiency. Other works attempt to decouple dynamics from different perspectives; for example, IFactor \citep{liu2023learning} decomposes latent variables based on their interaction with actions and rewards, while FlowDreamer \citep{guo2026flowdreamer} explicitly decomposes dynamics into the prediction of 3D scene flow and subsequent rendering.

Our decoupling paradigm offers a more flexible and general alternative to object-centric methods. By automatically separating primary and secondary dynamics at the feature level in a data-driven way, our framework naturally applies to complex scenarios such as deformable bodies and granular materials, where object segmentation is ill-defined.

\section{Implementation Details and Hyperparameters}
\label{app:hyperparams}

All our models are implemented based on PyTorch. The observation model $g_\phi$ uses the pre-trained DINOv2-ViT-S/14 model loaded from the HuggingFace Hub, with its weights kept frozen throughout all experiments. All our transition model components (History Fusion, Dynamic Localization, Main Predictor, LRM) are based on a standard ViT architecture. Detailed hyperparameters are provided in Table~\ref{tab:shared_hyperparams} and Table~\ref{tab:model_hyperparams}.

\begin{table}[h]
\centering
\caption{Shared Training Hyperparameters}
\label{tab:shared_hyperparams}
\begin{tabular}{lc}
\toprule
Hyperparameter & Value \\
\midrule
Optimizer & AdamW \\
Learning Rate & 7e-4 \\
Weight Decay & 0.01 \\
Batch Size & 64 \\
Training Epochs & 100 \\
Learning Rate Scheduler & Constant \\
\bottomrule
\end{tabular}
\end{table}

\begin{table}[h]
\centering
\caption{Model Architecture Hyperparameters}
\label{tab:model_hyperparams}
\begin{tabular}{lcccc}
\toprule
Module & Num Layers & Embed Dim & MLP Ratio \\
\midrule
Dynamic Localization Network & 6  & 192 & 4.0 \\
Sparse Main Predictor & 6  & 404 & 4.0 \\
History Fusion Module (Cross-Attention) & 1 & 404 & N/A \\
Low-Rank Correction Module (LRM) & 1 & 404 & N/A \\
\bottomrule
\end{tabular}
\end{table}

\section{CEM Planning Algorithm}
\label{app:cem}

In Section ~\ref{subsec:mpc} of the main text, we briefly introduced the use of CEM for planning. Algorithm~\ref{alg:cem} provides its detailed, step-by-step procedure. This algorithm is invoked at each decision step to search for the optimal action sequence.

\begin{algorithm}[h]
   \caption{CEM for Trajectory Planning}
   \label{alg:cem}
\begin{algorithmic}
   \STATE {\bfseries Input:} current latent state $z_t$, goal latent $z_g$, planning horizon $H$, number of iterations $K$, number of samples $N$, number of elites $E$.
   \STATE Initialize Gaussian distribution for action sequence $\mathcal{N}(\mu, \Sigma)$.
   \FOR{$i=1$ {\bfseries to} $K$}
   \STATE Sample $N$ action sequences $\{a_{t:t+H-1}^{(j)}\}_{j=1}^N$ from $\mathcal{N}(\mu, \Sigma)$.
   \STATE For each sequence $j$, perform rollout using DDP-WM to get future latents $\{\hat{z}_{t+1}^{(j)}, ..., \hat{z}_{t+H}^{(j)}\}$.
   \STATE Calculate cost $C_j = \mathcal{L}_{\text{MPC}}(\hat{z}_{t+H}^{(j)}, z_g)$ for each trajectory using the Sparse MPC Cost Mask.
   \STATE Select the top $E$ elite sequences with the lowest costs.
   \STATE Update $\mu$ and $\Sigma$ of the Gaussian distribution based on the elite set.
   \ENDFOR
   \STATE {\bfseries Output:} The mean of the final action distribution $\mu$ as the action for the first step.
\end{algorithmic}
\end{algorithm}

\section{Adaptive Sparse Size Mechanism}
\label{app:adaptive_size}

In Section~\ref{subsubsec:SPDP} of the main text, we mentioned an adaptive sizing strategy. Here we provide its detailed mechanism. This strategy aims to strike an optimal balance between computational efficiency and hardware utilization. We first set a hardware-friendly minimum sparse size, $k_{\text{min}}=32$. For a given batch, the actual sequence length fed to the primary predictor, $k_{\text{batch}}$, is dynamically set as:
\begin{equation}
k_{\text{batch}} = \max(k_{\text{min}}, \max_{i \in \text{batch}}(k'_i))
\end{equation}
where $k'_i$ is the number of changing regions actually detected for the $i$-th sample in the batch. If the number of changing regions for a sample, $k'_i$, is less than $k_{\text{batch}}$, we randomly sample $k_{\text{batch}} - k'_i$ patches from its static background regions to pad the input, ensuring that all input sequences in the batch have a regular length of $k_{\text{batch}}$.

This strategy offers a dual advantage: 1) In the vast majority of scenarios where $k'_i < k_{\text{min}}$, this padding ensures the regularity of the input tensors, thereby maximizing the utilization of optimizations in modern parallel computing hardware (like static computation graphs). 2) When encountering complex scenes with drastic changes that result in $k'_i > k_{\text{min}}$, the mechanism automatically and smoothly expands its computational capacity to handle these cases, without clipping important dynamic information due to a fixed sparsity budget.

\section{Pixel Error Calculation Details}
\label{app:pixel_error}

The pixel error is a metric we use to quantify the fidelity of the predicted visual features. 

For any given task, we use a single, frozen pixel decoder to reconstruct images from features. This decoder is typically the one co-trained with the baseline DINO-WM model.

It has a limited capacity. This is a deliberate choice to ensure that the reconstruction quality genuinely reflects the fidelity of the input features, rather than being an artifact of an overly powerful decoder that could overfit and mask underlying feature-level inaccuracies.

To compute the metric, we follow a consistent procedure: we randomly sample 400 episodes from the validation set, using a fixed random seed for fair comparison across all models. For each episode, we perform a 5-step open-loop rollout. The feature map from the final (fifth) predicted step is passed through the frozen decoder. The resulting image is compared pixel-by-pixel against the ground-truth image. We count the number of pixels where the absolute difference exceeds a predefined, task-agnostic threshold. The final ``Pixel Error" score is the average of this pixel count over the 400 samples.

\section{Environment and Dataset Details}
\label{app:env_details}

\subsection{PointMaze}
This environment, introduced by \citet{fu2021d4rldatasetsdeepdatadriven}, tasks a force-actuated 2-DoF ball to reach a target goal. Unlike simple kinematic models, the agent's dynamics incorporate realistic physical properties such as velocity, acceleration, and inertia. This requires the world model not merely to infer positions, but to model a dynamical system with physical momentum. For brevity, we refer to this task as Maze in our tables.

\subsection{Wall}
This is a custom 2D navigation environment that tests the model's spatial reasoning capabilities. The agent must plan within a non-convex space divided by a wall, finding and passing through a narrow door to navigate from one room to another.

\subsection{Push-T}
This environment, introduced by \citet{chi2024diffusionpolicyvisuomotorpolicy}, is a tabletop manipulation task that places high demands on physical reasoning. Success requires a precise understanding of the complex, contact-rich dynamics between the pusher agent and the T-shaped block, posing a significant challenge to the model's ability to capture the physics of rigid-body interaction.

\subsection{Rope Manipulation}
Introduced in \citet{zhang2024adaptigraphmaterialadaptivegraphbasedneural} and simulated in Nvidia Flex, this task involves interaction with a deformable body (a soft rope), a notoriously difficult problem in robotic manipulation. The model must learn and predict high-dimensional, non-rigid deformations induced by the robotic arm's actions.

\subsection{Granular Manipulation}
This environment uses the same simulation setup as Rope Manipulation but elevates the challenge to manipulating a multi-body system of approximately one hundred discrete particles. The model needs to understand the complex collective dynamics and collisional phenomena caused by pushing actions to gather the disordered particles into a desired shape.

\section{Training Strategy Details}
\label{app:training_strategy}

We adopt a stepwise, decoupled training strategy, primarily for its simplicity and robustness in implementation. A fully end-to-end joint training scheme would necessitate the introduction and careful tuning of multiple hyperparameter weights to balance the different loss functions for each module (e.g., localization, primary prediction, and LRM losses). Fine-tuning these weights is a complex and often brittle process that requires extensive, task-specific hyperparameter sweeps. Our stepwise approach circumvents this challenge entirely by decomposing the joint optimization problem into three simpler, independent sub-tasks. This makes the training process more stable and straightforward to reproduce. First, we train the Dynamic Localization Network on the complete offline dataset. If historical frames are used, the weights of the Historical Information Fusion Module are jointly trained during this stage. Subsequently, using the localization network, we train the Primary Dynamics Predictor. Its loss function is computed only on the foreground regions, aiming to minimize the MSE between the predicted foreground and ground-truth foreground features. Finally, with the first two modules, we train the Low-Rank Correction Module(LRM). Its loss function is computed only on the background regions, aiming to minimize the MSE between the compensated background and ground-truth background features.

\section{Qualitative Analysis of Open-Loop Rollouts}
\label{app:qualitative_rollouts}

This section provides a qualitative comparison of long-term open-loop predictions (5-step rollouts) between our DDP-WM and the dense baseline, DINO-WM, across three representative tasks: Push-T, Granular, and Rope. For a fair comparison, the same pixel decoder, originally co-trained with the DINO-WM baseline, is used to reconstruct images from the latent features of all models.

A visible and consistent pattern emerges across the tasks: predictions from the dense model, DINO-WM, can sometimes degrade over time, becoming blurry, distorted, and losing critical physical details. In contrast, our DDP-WM is more inclined to generate sharp, physically coherent, and high-fidelity predictions. We argue that this difference in prediction quality helps to explain the success of our method.

\subsection{Push-T Task}

\begin{figure}[h!]
    \centering
    \includegraphics[width=0.8\textwidth]{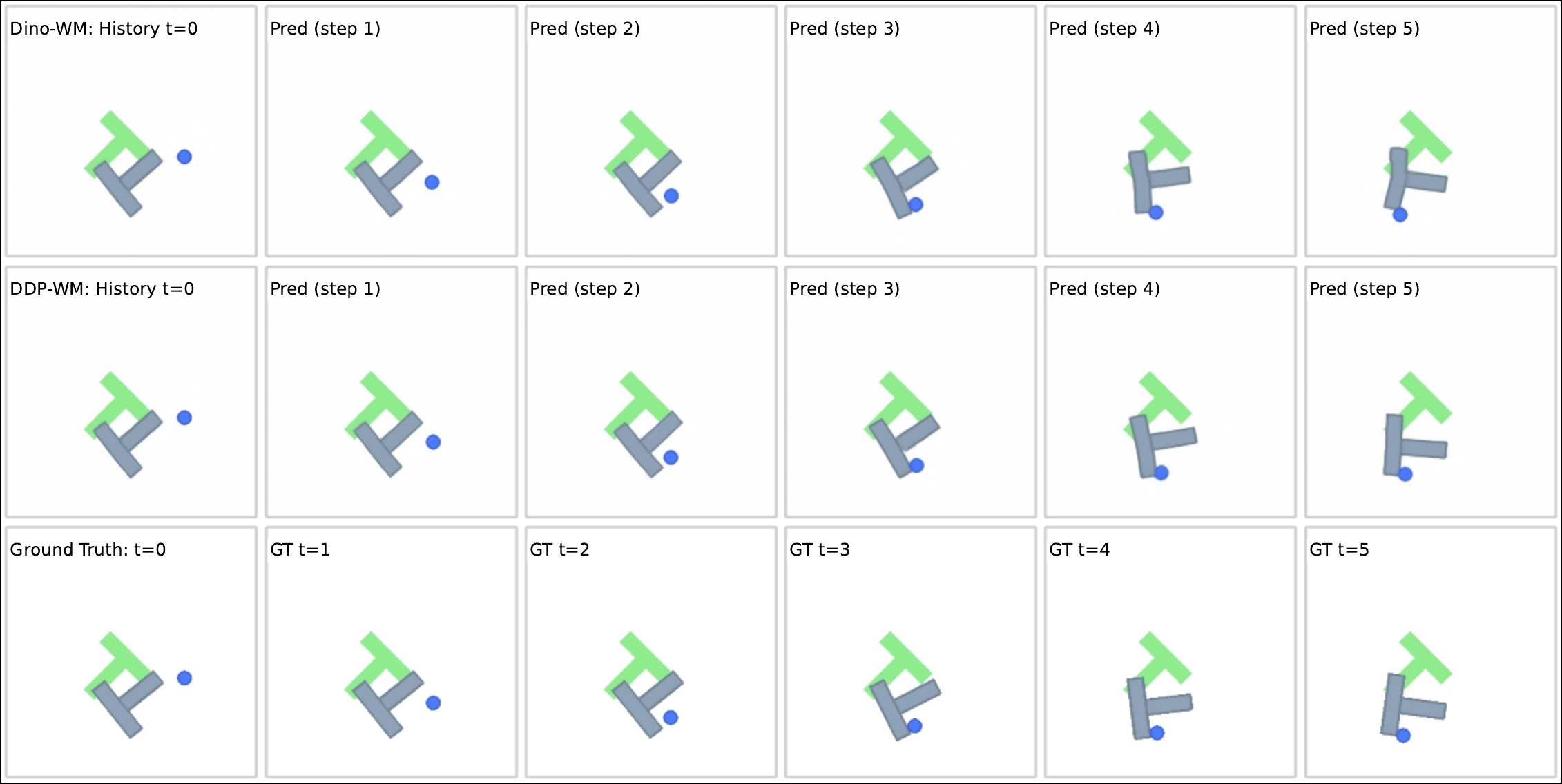}
    \includegraphics[width=0.8\textwidth]{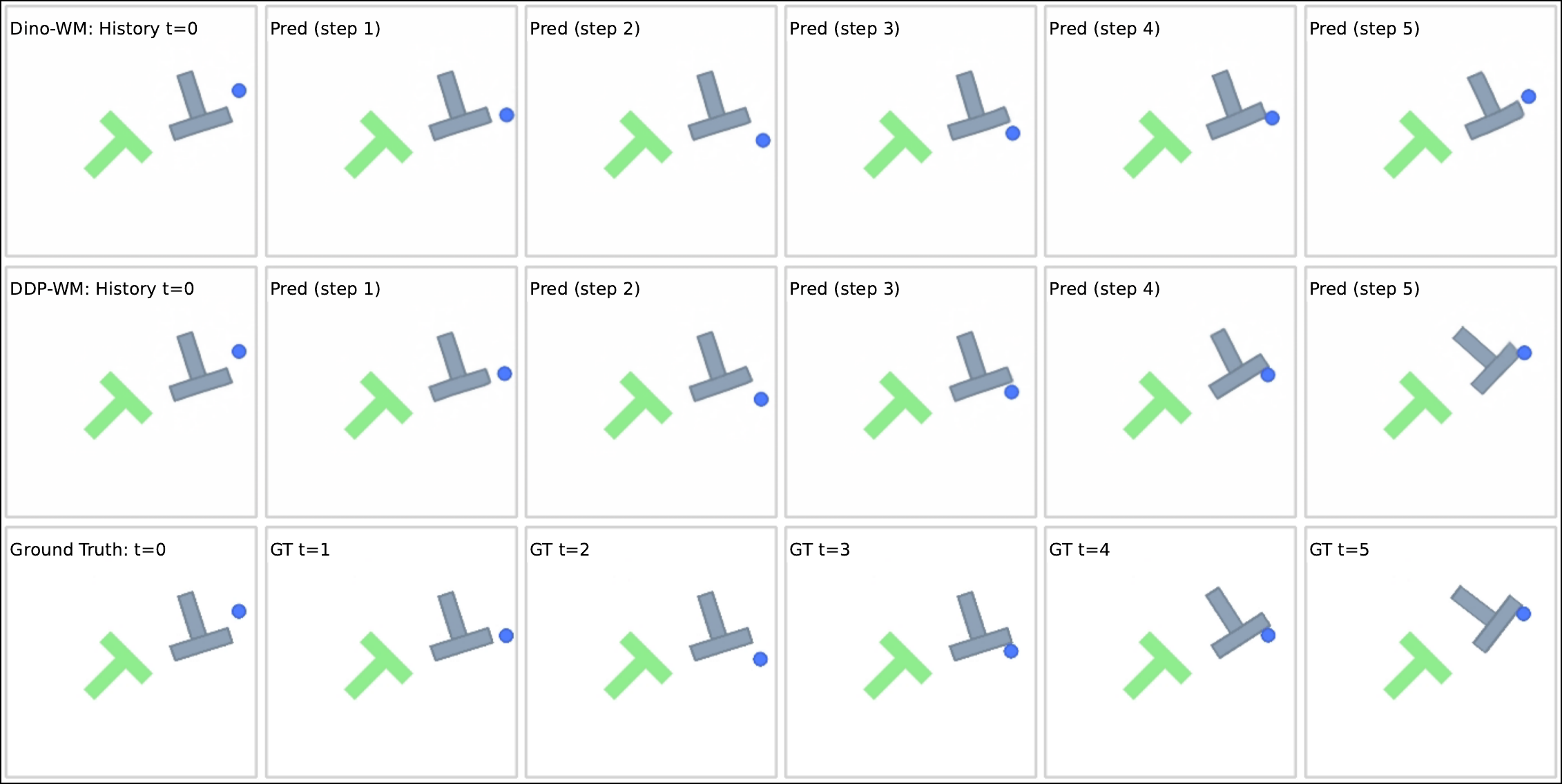}
    \includegraphics[width=0.8\textwidth]{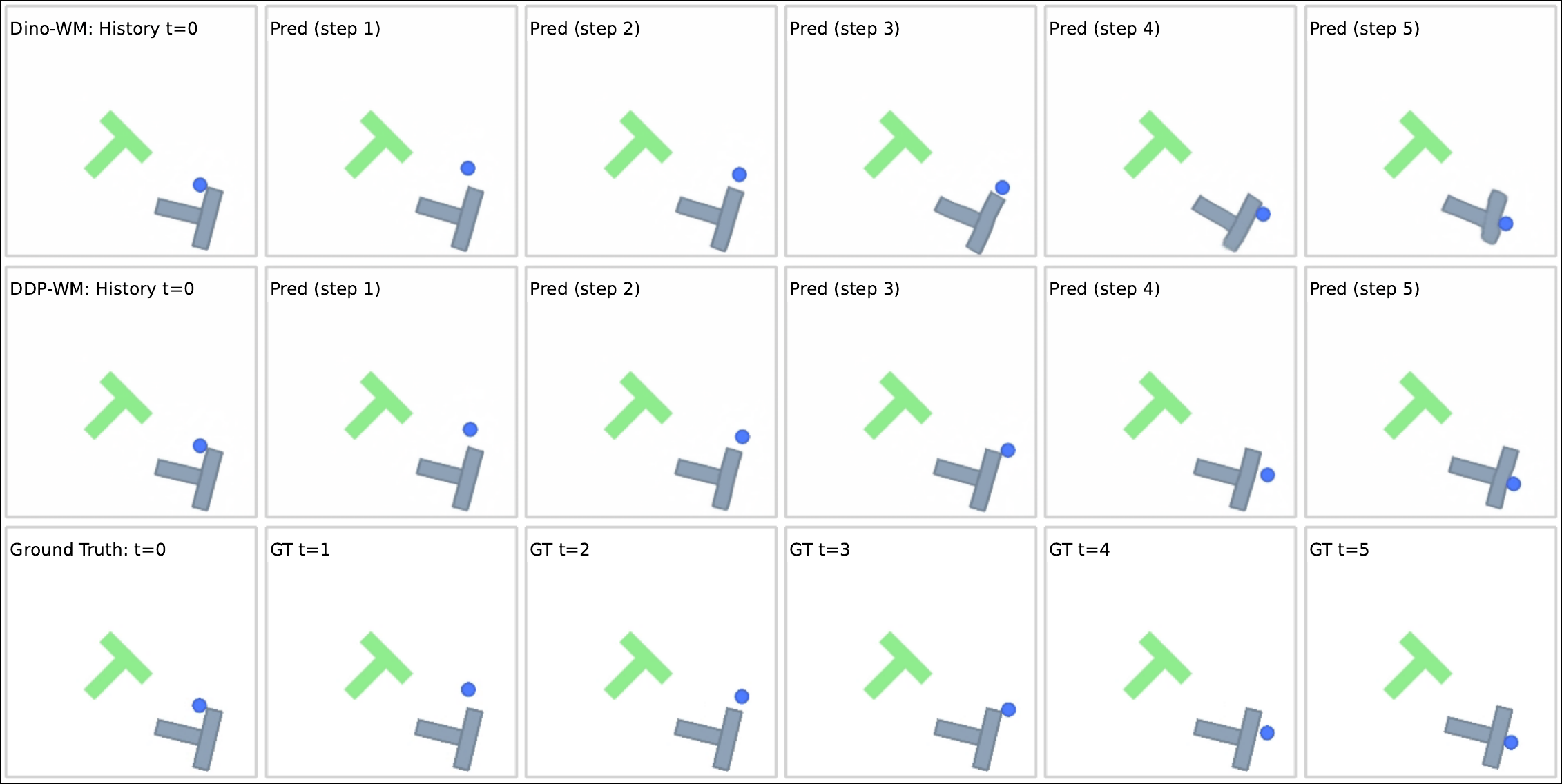}
    \caption{Open-loop rollouts on the Push-T task. From top to bottom: Sample 1, Sample 2, and Sample 3. For each sample, the rows show the results from DINO-WM (top), DDP-WM (Ours, middle), and the ground truth (bottom).}
    \label{fig:rollout_pusht_all}
    
\end{figure}

Figures~???\ref{fig:rollout_pusht_all} show some samples on the Push-T task. In DINO-WM's predictions, visual artifacts can be observed. For instance, the pushed T-block might exhibit feathering at its edges or soft-body-like distortions, which can compromise its true rigid-body characteristics. In contrast, our DDP-WM consistently maintains the block's sharp boundaries and correct rotational pose.

\subsection{Granular Manipulation Task}

\begin{figure}[h!]
    \centering
    \includegraphics[width=0.8\textwidth]{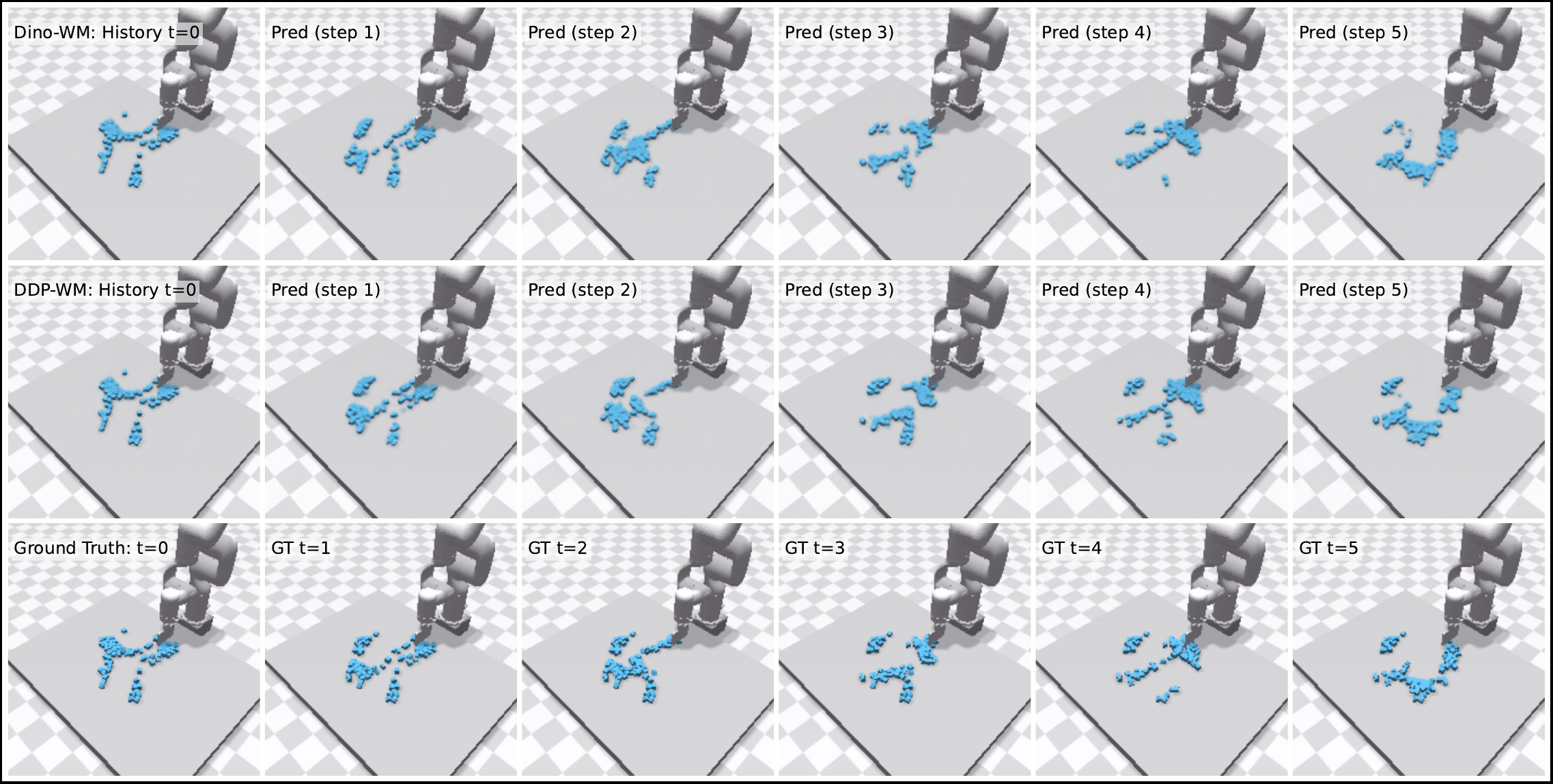}
    \includegraphics[width=0.8\textwidth]{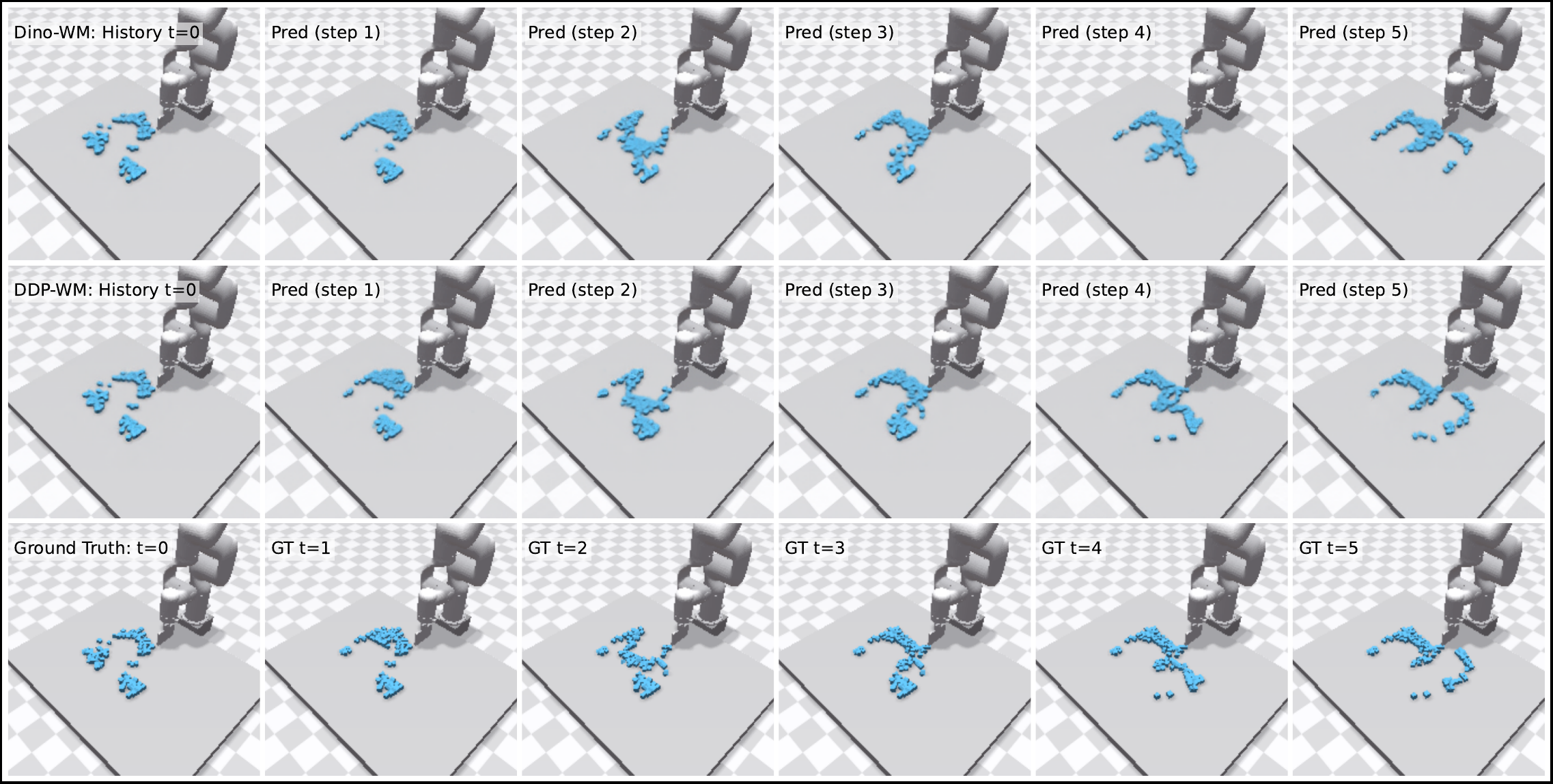}
    \includegraphics[width=0.8\textwidth]{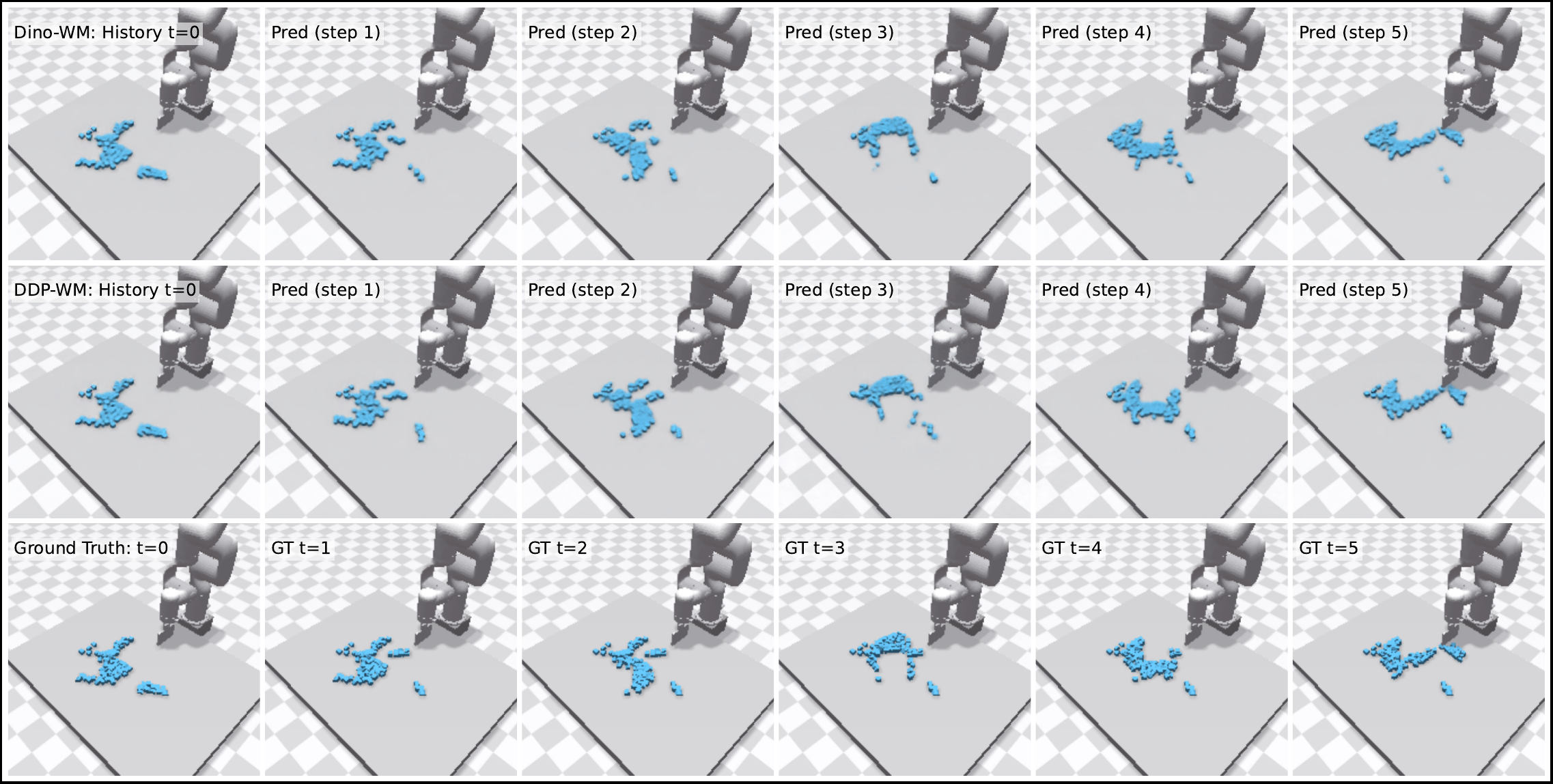}
    \caption{Open-loop rollouts on the Granular task. From top to bottom: Sample 1, Sample 2, and Sample 3. Similar to the previous figure, each sample shows results from DINO-WM (top row), DDP-WM (Ours, middle row), and the ground truth (bottom row).}
    \label{fig:rollout_granular_all}
\end{figure}

In the more challenging Granular task (Figure~??????\ref{fig:rollout_granular_all}), both models can sometimes misinterpret the concept of a multi-body system, incorrectly predicting the discrete particles as a continuous, amorphous blue "gel" or "cloud." However, our DDP-WM successfully preserves the discrete nature of the system in most cases, making predictions about the collective dynamics of the particles that are highly similar to the ground truth.

\subsection{Rope Manipulation Task}

\begin{figure}[h!]
    \centering
    \includegraphics[width=0.8\textwidth]{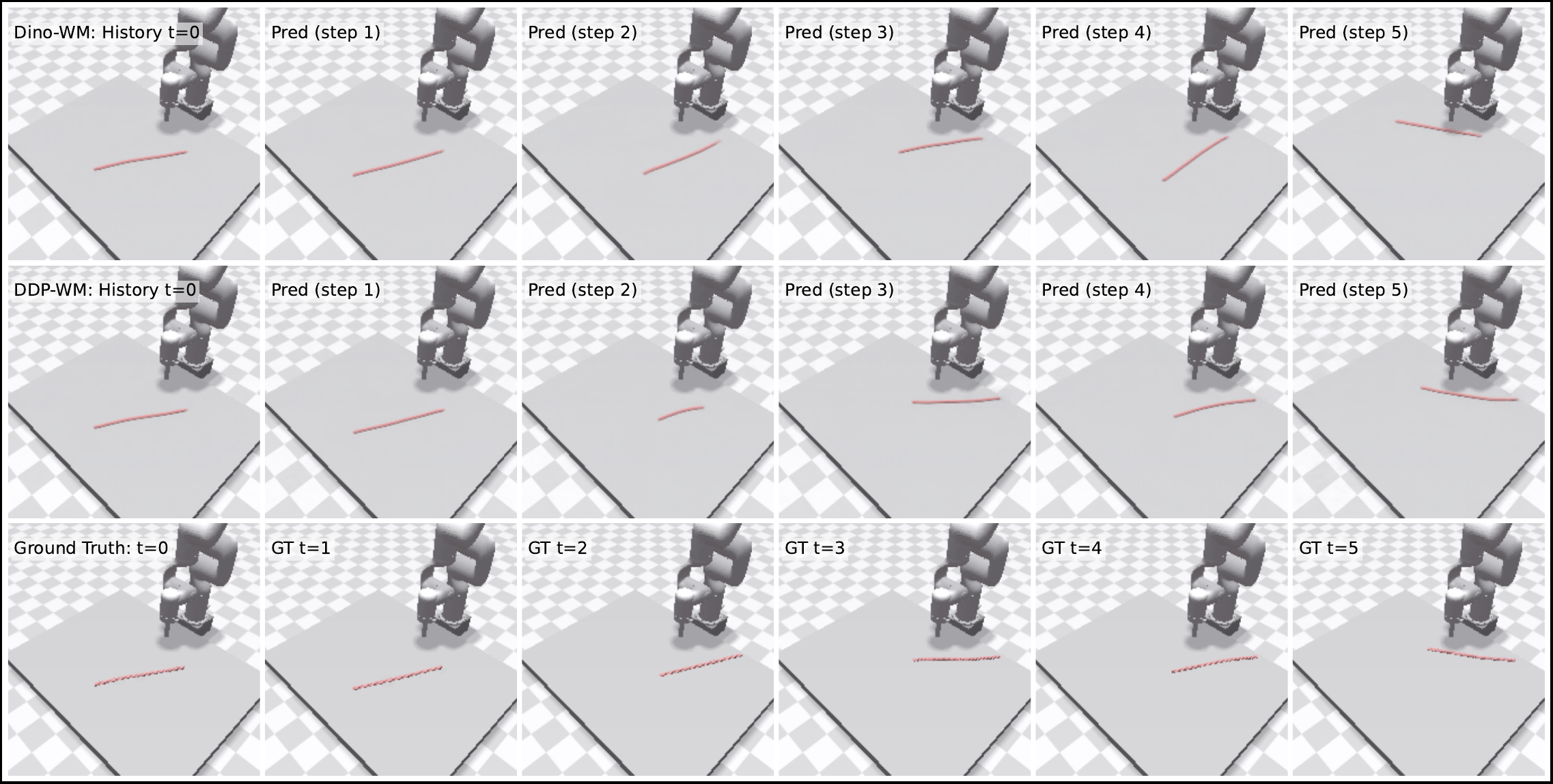}
    \includegraphics[width=0.8\textwidth]{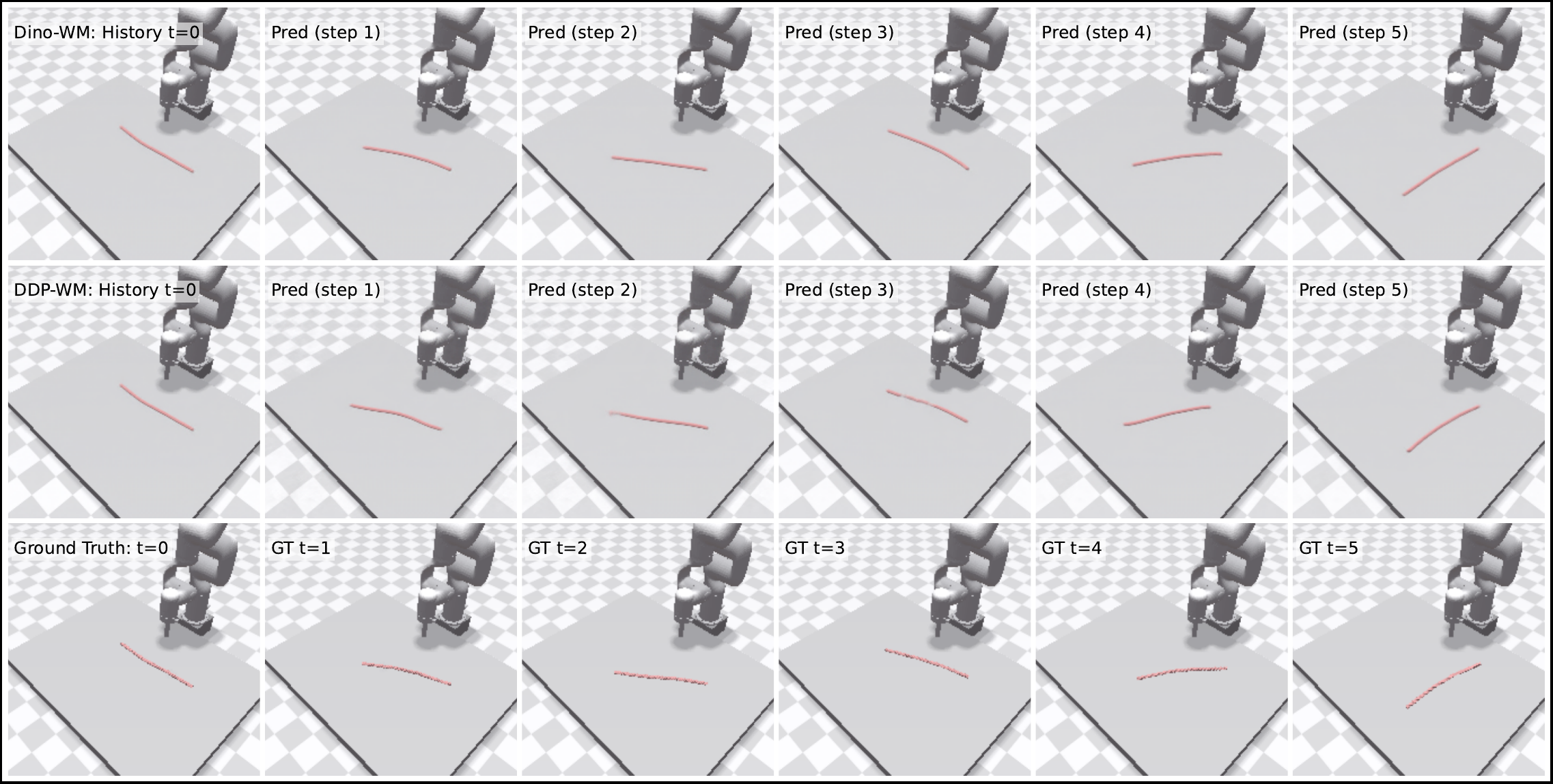}
    \includegraphics[width=0.8\textwidth]{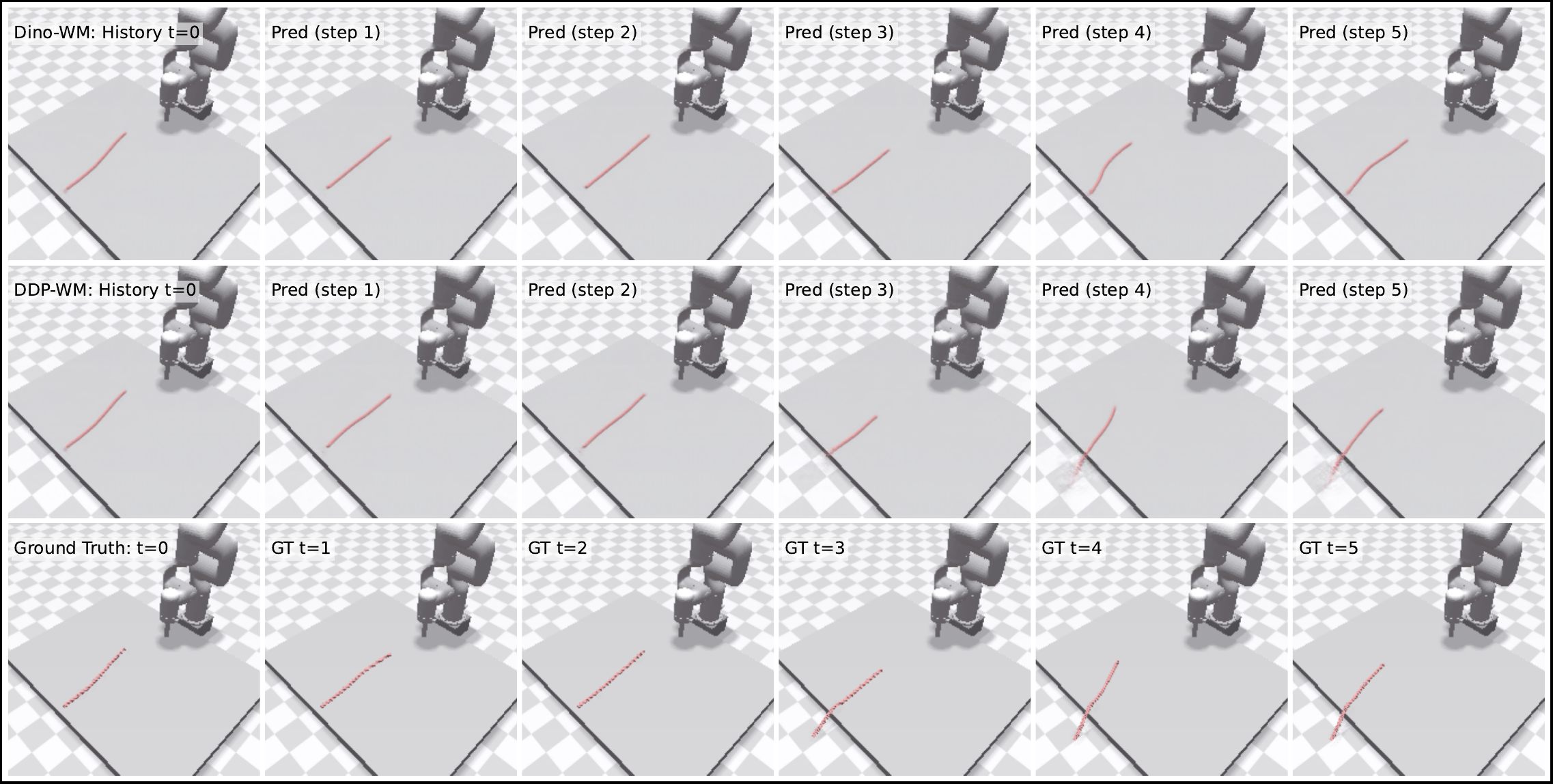}
    \caption{Open-loop rollouts on the Rope task. From top to bottom: Sample 1, Sample 2, and Sample 3. For each sample, the rows depict the rollouts from DINO-WM (top), DDP-WM (Ours, middle), and the ground truth (bottom).}
    \label{fig:rollout_rope_all}
\end{figure}

For the prediction of the deformable body (rope), as shown in Figure~\ref{fig:rollout_rope_all}, our DDP-WM consistently predicts the rope as a clear, continuous curve, accurately modeling its bending and twisting deformations.

\end{document}